\documentclass{article}

\usepackage{PRIMEarxiv}

\usepackage[utf8]{inputenc} 
\usepackage[T1]{fontenc}    
\usepackage{hyperref}       
\usepackage{url}            
\usepackage{booktabs}       
\usepackage{amsfonts}       
\usepackage{nicefrac}       
\usepackage{microtype}      
\usepackage{lipsum}
\usepackage{fancyhdr}       
\usepackage{graphicx}       
\graphicspath{{media/}}     

\usepackage[table]{xcolor}
\definecolor{azure}{rgb}{0.94,1.0,1.0} 
\usepackage{multirow}
\usepackage{natbib}
\usepackage[most]{tcolorbox}
\tcbset{
  keyfindingbox/.style={
    colback=gray!10, colframe=black!70,
    boxrule=0.5pt, arc=2pt, outer arc=2pt,
    left=6pt, right=6pt, top=4pt, bottom=4pt,
    fonttitle=\bfseries, before skip=10pt, after skip=10pt,
    after=\par\smallskip     
    
  }
}

\usepackage{rotating}
\usepackage{subcaption}
\usepackage{comment}
\usepackage{longtable}
\usepackage[flushleft]{threeparttable}
\usepackage{enumitem}

\renewcommand{\arraystretch}{1.3}
\definecolor{lightblue}{RGB}{221,235,247}

\title{Mind Reading or Misreading? LLMs on the Big Five Personality Test
}

\author{
  Francesco Di Cursi, Chiara Boldrini, Marco Conti, Andrea Passarella \\
  IIT-CNR \\
  Pisa\\
  \texttt{\{francesco.dicursi, chiara.boldrini, marco.conti, andrea.passarella\}@iit.cnr.it} \\
}

\begin{document}
\maketitle

\begin{abstract}
We evaluate large language models (LLMs) for automatic personality prediction from text under the binary Five Factor Model (BIG5). Five models---including GPT-4 and lightweight open-source alternatives---are tested across three heterogeneous datasets (Essays, MyPersonality, Pandora) and two prompting strategies (minimal vs. enriched with linguistic and psychological cues). Enriched prompts reduce invalid outputs and improve class balance, but also introduce a systematic bias toward predicting trait presence. Performance varies substantially: Openness and Agreeableness are relatively easier to detect, while Extraversion and Neuroticism remain challenging. Although open-source models sometimes approach GPT-4 and prior benchmarks, no configuration yields consistently reliable predictions in zero-shot binary settings. Moreover, aggregate metrics such as accuracy and macro-F1 mask significant asymmetries, with per-class recall offering clearer diagnostic value. These findings show that current out-of-the-box LLMs are not yet suitable for APPT, and that careful coordination of prompt design, trait framing, and evaluation metrics is essential for interpretable results.
\end{abstract}

\keywords{Automatic personality prediction \and Binary Big Five \and Large language models \and Prompt enrichment \and Class-wise evaluation}

\section{Introduction}
\label{sec:intro}

Automatic Prediction of Personality from Text (APPT) concerns the task of inferring an author's psychological traits solely from their written language. This challenge lies at the intersection of psychology, computational linguistics, and machine learning, and remains unresolved despite the remarkable progress in NLP of large language models (LLMs). While models such as GPT-4 and LLaMA have demonstrated impressive abilities in general-purpose language understanding, it is unclear whether they can reliably capture psychologically meaningful constructs, particularly in zero-shot settings. Focusing on zero-shot use is crucial, as it reflects how LLMs are often deployed in real-world applications where labeled data are scarce, and it establishes a lower bound of their capabilities before considering costly fine-tuning.

In this work, we focus on binary classification within the Five Factor Model~\cite{costa1992normal} (also known as the Big Five or BIG5 or OCEAN), a well-established framework in personality psychology.\footnote{Further details on the theoretical foundations of APPT and BIG5 can be found in the Lexical Hypothesis~\cite{galton1884measurement, allport1936trait, goldberg2013alternative} and Brunswik's Lens Model~\cite{brunswik1955perception}.} The model defines five core traits: Openness (OPN; curiosity, imagination), Conscientiousness (CON; organization, self-discipline), Extraversion (EXT; sociability, assertiveness), Agreeableness (AGR; empathy, cooperativeness), and Neuroticism (NEU; emotional instability). For each trait, LLMs are asked to predict whether it is present (1) or absent (0) in a given text, without prior annotations or fine-tuning. This binary setup, widely adopted in APPT research, offers interpretability and simplicity but also exacerbates challenges such as class imbalance and weakly expressed signals. Despite its reductive nature, the binary framing remains a valuable benchmark: it enables direct comparison with pre-LLM literature and magnifies the methodological challenges that more complex labelings (e.g., continuous, trichotomous) often conceal.

We evaluate five LLMs: GPT-4 and four lightweight open-source models. The inclusion of small, openly available models reflects realistic resource constraints in applied APPT and ensures transparency and reproducibility. Moreover, testing whether compact models can approximate proprietary systems like GPT-4 informs both scientific understanding and practical deployment in low-resource settings. We evaluate these models across three linguistically and contextually diverse datasets: stream-of-consciousness Essays~\cite{pennebaker1999linguistic}, aggregated Facebook updates from MyPersonality~\cite{kosinski2013private}, and Reddit posts from Pandora~\cite{gjurkovic2020pandora}. These datasets differ in language style, length, author intent, and trait salience, enabling us to probe model robustness under heterogeneous conditions. Two prompting strategies\footnote{Prompting refers to the use of natural language instructions to elicit model predictions, as opposed to fine-tuning the model on labeled data.} are compared: (i) a minimal, task-oriented prompt, and (ii) a linguistically and psychologically enriched prompt incorporating high- and low-level trait descriptors. Model performance is assessed using complete classification reports, including accuracy, per-class F1, precision, and recall, to highlight asymmetries between trait presence and absence predictions.

Our findings show that enriched prompts generally improve output validity, particularly for models that struggle with minimal instructions, and can partially balance class predictions. However, this improvement comes at a cost: enriched prompts induce a consistent positive bias toward predicting trait presence. We also observe heterogeneous failure modes: GPT-4 often refuses or evades predictions due to safety constraints, whereas models such as Phi3 tend to produce invalid or incoherent outputs. Furthermore, some traits (notably Openness and Agreeableness) are more consistently identifiable, while Extraversion and Neuroticism remain elusive even under enriched prompting. Crucially, no model in our study surpasses the strongest pre-LLM systems. In a single configuration, one open-source model approximates prior state-of-the-art performance, but only under enriched prompting, underscoring the need for careful coordination between prompt design and trait framing.

Taken together, our design choices reflect a deliberate strategy: we adopt the binary framing, the zero-shot setting, and lightweight open-source models not as constraints but as stress tests. The binary setup provides interpretability and continuity with prior APPT literature, while the zero-shot condition probes the out-of-the-box capabilities of LLMs in scenarios where labeled data are scarce. Evaluating lightweight open-source models alongside GPT-4 ensures transparency, reproducibility, and relevance to practical deployments where computational resources are limited. This philosophy allows us to highlight methodological challenges and opportunities under the most accessible and broadly applicable conditions.

Against this background, our study is guided by the following research questions:
\begin{enumerate}[label=\textbf{RQ\arabic*}, leftmargin=*, align=left]
    \item To what extent can current LLMs predict Big Five personality traits from text under a binary framing and zero-shot conditions?
    \item How does prompt design (minimal versus enriched with linguistic and psychological cues) affect output validity, class balance, and predictive performance?
    \item Which evaluation metrics provide the most reliable insights into LLM behavior on APPT, and what methodological lessons can be drawn for future studies?
\end{enumerate}

This study makes six key contributions: (i) a multi-model, multi-dataset evaluation of LLMs on binary personality prediction across five traits; (ii) a systematic comparison of minimal and enriched prompting strategies across 135 experiments; (iii) evidence that enriched prompts reduce invalid outputs but induce systematic positive bias; (iv) an analysis of heterogeneous model failures, from task misunderstandings and refusals due to lack of context or out of scope task in smaller models to safety-driven refusals in GPT-4; (v) a critical reassessment of evaluation metrics, showing that per-class recall reveals behavioral asymmetries that accuracy and macro-F1 conceal, while precision largely mirrors trait distribution; and (vi) the observation that, in one constrained case, an open-source model approximates pre-LLM state-of-the-art results---not as a sign of progress, but as a benchmark for zero-shot prompting.

\textit{Ethical considerations.} Automatic personality prediction  carries risks of unfair profiling, privacy violations, manipulation (e.g., in advertising or politics), and misuse in high-stakes domains such as hiring, healthcare, or justice. It may also reinforce cultural biases and disadvantage marginalized groups. Our study does not promote such uses; rather, by exposing the unreliability and biases of current LLM-based approaches, we aim to discourage premature or harmful applications. Were such systems to be implemented in practice, risks could be mitigated by giving individuals full control over the analysis—for example, by allowing APPT tools to run locally on a user’s own data without disclosing it to external servers.

The remainder of the paper is organized as follows. Section~\ref{sec:related_works} reviews related work on APPT with LLMs. Section~\ref{sec:methodology} details the datasets, models, prompts, and evaluation setup. Section~\ref{sec:results} presents experimental results and key observations. Section~\ref{sec:conclusion} concludes with methodological lessons and future directions.

\section{Related Works}
\label{sec:related_works}

Recent research has examined the ability of large language models (LLMs), such as GPT-3, GPT-4, and LLaMA, to infer Big Five personality traits from text via prompting. These studies differ in task framing, evaluation strategies, and labeling schemes (binary, continuous, or trichotomous). Below we present a thematic review, with particular attention to evaluation metrics and the reporting of label distributions and text statistics.

\citet{peters2024large} recruited 600 US participants from Prolific Academic, administered the BFI-2 questionnaire\footnote{The BFI-2 questionnaire comprises 60 items (12 per trait). Shorter versions include the BFI-2-S (30 items) and BFI-10 (10 items).}, and evaluated GPT-4's ability to predict continuous trait scores from different types of user–LLM interactions. Users could interact in either short conversations (User, Conversation) or unconstrained usage (User, Unconstrained Use). The LLM, in turn, adopted different roles: assessing personality (Bot, Assessment), getting acquainted with the user (Bot, Acquaintance), or serving as a neutral assistant (Bot, Assistant). Results, measured with Pearson correlation, showed the highest scores in the Assessment setting (mean correlations $\approx$0.44–0.45), while the Assistant condition performed worst. Although correlations are high (0.3–0.4), trait distributions and text statistics were not provided.

A comparable study by~\citet{piastra2025emergent} examined zero-shot GPT-4 predictions on the Essays~\cite{pennebaker1999linguistic} and PAN15~\cite{rangel2015overview} datasets\footnote{PAN15 is available upon request at \url{https://zenodo.org/records/3745945}; Essays can be accessed at \url{https://github.com/SenticNet/personality-detection/blob/master/essays.csv}.}. Instead of role-playing user–bot interactions, the study varied prompts through role-play strategies. The baseline was a ``social psychology expert'' role, extended with enriched prompts including BFI items and Goldberg adjectives~\cite{goldberg1992development}. For PAN15, the impact of text length was also analyzed. Best correlations range between 0.25 and 0.29. Interestingly, explicitly requesting confidence from GPT-4 produced unreliable results and degraded performance. As in \citet{peters2024large}, label distributions and text statistics were not reported.

\citet{derner2024can} conducted a controlled experiment with 155 Czech participants, each completing the BFI-44 questionnaire twice (self- and partner-reports) and producing four short texts\footnote{Texts included a cover letter, vacation letter, complaint letter, and apology letter, each $\sim$200 words.}. Two external raters also provided trait predictions. Labels were trichotomous (low, medium, high). Results showed that human raters and partners consistently outperformed GPT-4. A strong positivity bias emerged: GPT-4 frequently predicted ``high'' and rarely ``low''. The best results were obtained when prompts combined task and trait descriptions. Notably, this work reported label distributions and class-wise hit rates, highlighting how aggregate F1 scores can be misleading.

\citet{yan2024predicting} investigated LLaMA3-8B and Qwen1.5-110B on Chinese counseling dialogues, predicting continuous scores using Pearson correlation and MAE. Roles included client, counselor, and observer, with BFI-2 items provided as context for LLaMA3-8B. Findings indicated that at least 30\% of a dialogue was required for stable predictions, and that fine-tuned LLaMA3-8B (via DPO and SFT) outperformed Qwen-110B. Performance scaled with model size. Reported correlations range between 0.3 and 0.5, though no text statistics or distributions were reported.

In contrast, \citet{zhu2025investigating} evaluated GPT-4 and GPT-4-mini on 15-minute interviews from 102 participants who also completed the BFI-10. Prompts were enriched with trait facets from \citet{matthews2003personality}. Results, assessed via Pearson correlation and mean difference, were mostly null or negative, with only Agreeableness showing a moderate correlation (0.25). GPT-4 and GPT-4-mini correlated strongly with each other ($\geq$0.50), suggesting consistent but poor performance. Trait distributions were not reported, and reliance on the brief BFI-10 questionnaire reduced interpretability, as short inventories generally yield low resolution and weaker results.

\citet{ji2023chatgpt} tested GPT-3.5 on binary classification with the PAN15 and Essays datasets, using median splits for labels. Baselines included ROBERTA and RNN classifiers with fixed hyperparameters. Several prompting strategies were tested: zero-shot, chain-of-thought, one-shot, and a ``level-oriented'' formulation inspired by hierarchical NLP classifiers. Due to input constraints\footnote{Multi-shot prompting risks exceeding token limits or producing invalid inputs due to a single problematic text.}, only one-shot examples were feasible. Accuracy was the sole evaluation metric, and reported accuracies of their framework are between 0.5 and 0.6 in the case of the Essays dataset, and between 0.6 and 0.7 in in the case of PAN15 dataset. However, their reported SOTA values, especially for Essays ($\sim$0.8), appear inflated\footnote{\citet{zhu2022lexical}, from which \citet{ji2023chatgpt} draws SOTA values, reported $\sim$0.8 accuracy for HPMN (BERT) on balanced Essays but without class-wise metrics. For example, a model predicting 400/500 ``high'' correctly but only 200/500 ``low'' correctly yields 80\% accuracy, yet the negative-class F1 would fall to $\sim$0.3. Such imbalances are hidden without class-wise reporting.} underscoring the importance of class-wise evaluation. \citet{ji2023chatgpt} also found that chain-of-thought and level-oriented prompting improved results.

\citet{ganesan2023systematic} studied binary classification using GPT-3 on MyPersonality (users with $\geq$20 posts). Multiple prompt templates were tested, including textbook trait descriptions, word lists, and questionnaire items. The authors do not specify which questionnaire was used. Their approach parallels ours, though they applied separate prompts rather than combining descriptions. Baselines included GPT-3 with basic prompts and a regression model from \citet{park2015automatic}. Both accuracy and F1 were evaluated. Reported F1 ranged from $\sim$0.3 to $\sim$0.56, with regression outperforming GPT-3 except in some questionnaire-based cases (F1 $\geq$0.5). Authors noted that trichotomous labels degraded performance. A positivity bias was again observed, with GPT-3 often predicting socially desirable traits (e.g., high Openness, low Neuroticism).

\citet{amin2023will} evaluated GPT (the authors refer to the model generically as ChatGPT, stating that they accessed it on 30 January 2023) on the binary BIG5 subset of the Impression dataset (access to the Impression dataset requires consent, and the URL is currently inactive). They used minimal prompts to predict binary labels for each trait, treating the task as multiclass classification. For comparison, they tested traditional pipelines with different encodings (subwords, words) and embeddings (RoBERTa, Word2Vec, TF-IDF) combined with classifiers (MLP, SVM). Unlike \citet{ji2023chatgpt} and \citet{ganesan2023systematic}, they explicitly addressed class imbalance: labels were balanced except for Openness and Agreeableness (skewed 2:1). They reported both accuracy and Unweighted Average Recall (UAR)\footnote{Accuracy is class-agnostic and remains high if only the majority class is predicted, whereas UAR penalizes such imbalances.}. Results showed that RoBERTa embeddings outperformed alternatives (accuracy 0.62–0.67, UAR $\sim$0.6), while GPT performed poorly across both metrics.

In summary, prior work spans continuous, trichotomous, and binary labeling formats, with heterogeneous evaluation protocols. Crucially, works also span different data, with works collecting data through small experiments, and other leveraging already existing datasets. Continuous prediction typically relies on Pearson correlation and MAE, while binary classification emphasizes accuracy and F1. Few studies report class-wise metrics such as UAR or per-class hit rates, limiting interpretability. Trait and text distributions are rarely disclosed, complicating assessments of model robustness under imbalance or bias. Several works report positivity bias, where models overpredict trait presence. Overall, evaluation quality remains uneven, and results vary in interpretability and reliability across studies. 

To facilitate comparison with pre-LLM results, we report here the best SOTA values for binary predictions on the datasets considered. For Essays, SOTA binary trait predictions generally fall in the range of 0.55–0.65 across different works summarized in~\cite{ramezani2022text}.  Slightly higher values reported in~\cite{tandera2017personality} for MyPersonality. For Pandora (a relatively new dataset provided with continuous labels), prior studies have focused on regression rather than binary classification, so no direct SOTA values are available for the latter. Although some works report higher results, these are exceptions and usually lack class-wise reporting; therefore, we take the recurring value of around 0.6 as a representative SOTA benchmark. While SOTA results in APPT are not always fully comparable due to differences in data preprocessing, experimental setups, and evaluation metrics, we believe that these pre-LLM numbers provide a useful reference point for interpreting the out-of-the-box performance of LLMs, which is the focus of this work.

\section{Methodology and experimental setup}
\label{sec:methodology}

\subsection{Datasets}
\label{sec:datasets}

We employ three widely used datasets covering different contexts:  
(i) \textbf{Essays} (long stream-of-consciousness texts written under psychological guidance, with binary labels),  
(ii) \textbf{MyPersonality} (Facebook status updates, with both continuous and binary labels), and  
(iii) \textbf{Pandora} (Reddit posts, originally labeled on a discrete scale).  

While Essays and MyPersonality already provide binary labels, Pandora requires binarization. To achieve this, we use MyPersonality as reference, given its heavily right-skewed score distributions and derived binary labels. Since Pandora scores are integers in the 1–100 range and MyPersonality scores are continuous in the 1–5 range, we interpolate Pandora values to match MyPersonality’s scale. The threshold is then set to the highest MyPersonality value still labeled as 0 (i.e., not exhibiting the trait), ensuring that Pandora’s positive class corresponds to texts with strong linguistic cues of trait presence. Traditional binarization often relies on mean or median splits. Here, we adopt a thresholding strategy that maximizes the contrast between negative and positive labels. In MyPersonality, the binary split occurs at the peak of a skewed distribution, meaning that many ``negative'' labels actually correspond to weakly positive cases. In Pandora, by contrast, we define positives only for high-scoring cases, yielding fewer but stronger signals. This produces an imbalanced dataset but ensures linguistic clarity in the positive class.

For each dataset, Table~\ref{tab:summary} reports descriptive statistics: number of texts and users, mean and standard deviation of word and character counts, and the distribution of Big Five binary traits.

\begin{table*}[!t]
    \centering
      \begin{threeparttable} 

    \resizebox{\linewidth}{!}{
    \begin{tabular}{lcccccc ccccc}
        \toprule
        \textbf{Dataset} & \textbf{\# Texts} & \textbf{\# Users} 
        & \multicolumn{2}{c}{\textbf{Text Length [Characters]}} 
        & \multicolumn{2}{c}{\textbf{Text Length [Words]}} 
        & \textbf{OPN (1/0)} & \textbf{CON (1/0)} & \textbf{EXT (1/0)} & \textbf{AGR (1/0)} & \textbf{NEU (1/0)} \\
        \cmidrule(lr){4-5} \cmidrule(lr){6-7}
        & & & \scriptsize{\textbf{Median}} & \scriptsize{\textbf{SD}} 
        & \scriptsize{\textbf{Median}} & \scriptsize{\textbf{SD}} 
        & & & & & \\
        \midrule
        Essays & 2467 & 2467 & 3164.0  & 1287.02  & 726.0  & 306.72  & 1271 / 1196  & 1253 / 1214  & 1276 / 1191  & 1310 / 1157  & 1233 / 1234 \\
        MyPersonality & 255 & 255  & 1739.5 & 3964.53  & 317.5  & 712.79  & 176 / 74  & 130 / 120  & 96 / 154  & 134 / 116  & 99 / 151 \\
        Pandora & 14221 & 1608 & 131.0  & 438.96  & 25.0  & 77.86  & 6421 / 7800  & 2512 / 11709  & 2902 / 11319  & 2667 / 11554  & 5334 / 8887 \\
        \bottomrule
    \end{tabular}
    }
    \caption{Summary statistics and class distributions for binary personality traits. For each trait, 1 indicates presence and 0 indicates absence, and the number of text with those traits are reported as \textit{num1/num0}, where \textit{num1} is the number of texts with the trait and \textit{num0} the number without. Note that in the MyPersonality dataset, posts have been aggregated by author, resulting in a single merged text per individual.}
    \label{tab:summary}
    \end{threeparttable}
\end{table*}

\subsection{Models}

We evaluate \texttt{GPT-4} and selected open-source LLMs accessed via Ollama\footnote{Ollama is an open-source framework for downloading and running LLMs locally. See the GitHub repository at: \url{https://github.com/ollama/ollama}}:
\begin{itemize}
    \item \texttt{GPT-4} ($\sim$1T)\footnote{This is an estimate\cite{Strickland2024WhatIsGenerativeAI}, given that OpenAI did not release the number of parameters.}
    \item \texttt{phi3:latest} ($\sim$3.8B parameters)  
    \item \texttt{gemma:7b-instruct} (7B parameters)  
    \item \texttt{llama3.1:8b} (8B parameters)  
    \item \texttt{mistral:instruct} (7B parameters)  
\end{itemize}
Other models (e.g., Zephyr, Falcon, Minerva) were excluded as they consistently failed to return valid binary outputs. Grok-1 was also excluded due to its excessive size (100B parameters, $\sim$120GB), which is disproportionate compared to the lightweight models under study ($\sim$8B parameters, $\sim$2GB).  

\subsection{Prompting Strategies}

We define two prompting strategies:

\paragraph{Simple prompt.}  
A minimal task description instructs the model to return a binary prediction. The template is:

\begin{flushleft}\scriptsize{
\texttt{simple\_system\_prompt = f"""You are a psychological text classifier. Your job is to analyze a written text and decide if the BIG5 trait **\{TRAIT\}** is expressed. You are evaluating the content of the text, not the person.\footnote{Focusing on the text rather than the individual reduces refusal rates, since many models decline tasks framed as psychological assessment.} If the trait is present in the text, respond with 1. If not, respond with 0. You must always respond with a single digit: 0 or 1. Do not explain or add anything else."""}}
\end{flushleft}
\normalsize
All predictions are generated with temperature $=0$ for reproducibility and a maximum output length of 20 tokens\footnote{Ollama does not allow restricting models to output only specific tokens (e.g., ``0'' or ``1''). Models such as Phi3 sometimes embed the correct label inside a longer sentence. We allow up to 20 tokens in such cases, although only the strict single-digit outputs --- or those starting with a single digit --- are used for evaluation. Verbose predictions are stored for future qualitative analysis.}.

\paragraph{Complex prompt.}  
A linguistically and psychologically rich prompt integrates three sources:  
(i) high-level trait descriptions from~\citet{costa1992normal};  
(ii) facet-level positive/negative cues from the IPIP inventory\footnote{\url{https://ipip.ori.org/newBigFive5broadKey.htm}};  
(iii) adjective lists from \citet{goldberg2013alternative}.  
All these descriptions are provided in the appendix.
This richer context is designed to guide the model beyond surface-level cues. The following is the general structure of the complex prompt:

\begin{flushleft}\scriptsize
\texttt{complex\_prompt = f"""\string
You are an expert in Automatic Personality Prediction using the Five Factor Model (BIG5). Estimate whether the trait **\{TRAIT\}** is reflected in the style of the text, where "1" means present and "0" absent.\string\\
Base your judgment on both explicit and implicit cues --- including tone, content, emotional framing, and vocabulary.\string\\
\string\\
---\string\\
\#\#\# Trait Overview\string\\
**\{TRAIT\_DESCRIPTION\}**\string\\
\string\\
---\string\\
\#\#\# Facet-level Insight\string\\
If the text reflects these characteristics $\rightarrow$ output "1": \{LONG\_POSITIVE\_DESCRIPTIONS\}\string\\
If the text reflects these characteristics $\rightarrow$ output "0": \{LONG\_NEGATIVE\_DESCRIPTIONS\}\string\\
\string\\
---\string\\
\#\#\# Adjective Insight\string\\
Traits associated with output "1": \{POSITIVE\_ADJECTIVES\}\string\\
Traits associated with output "0": \{NEGATIVE\_ADJECTIVES\}\string\\
"""}
\end{flushleft}
\normalsize

\subsection{GPT-4 Setup}

For GPT-4, we used only the simple prompt due to time and cost constraints --- being the explanation out of the scope of this work --- via the OpenAI API. Experiments with GPT-4 were conducted using the following template:

\begin{flushleft}\scriptsize
\texttt{gpt4\_simple\_system\_prompt = f"Classify the text based on the trait of \{TRAIT\} from the OCEAN model. Use a scale where 1 represents high or moderate levels of \{TRAIT\}, and 0 represents low levels. Rely on semantic and contextual cues. Output format: a single character, either 0 or 1."}
\end{flushleft}
\normalsize
Minor adjustments (i.e., stricter output rules) were later introduced for open-source models to ensure consistency and avoid refusals or verbose outputs, but GPT-4 predictions were preserved as they contained no such errors.

\subsection{Evaluation}
\vspace{-4pt}

We assess model outputs using the full classification report for binary prediction: overall accuracy, average F1 (i.e., by averaging between the F1 of the two classes for a given experiment), and per-class precision, recall, and F1. This allows us to capture asymmetries between positive and negative predictions that are often obscured by aggregate measures.

\begin{table*}[ht]
\centering
\small
\setlength{\tabcolsep}{3pt}
\renewcommand{\arraystretch}{0.95}
\begin{tabular}{lllrrrrr}
\toprule
\textbf{Dataset} & \textbf{Prompt} & \textbf{Model} & \textbf{O} & \textbf{C} & \textbf{E} & \textbf{A} & \textbf{N} \\
\midrule
\rowcolor{white}
MyPersonality & Simple & llama3.1:8b & 3 (1.2\%) & 3 (1.2\%) & 3 (1.2\%) & 3 (1.2\%) & 3 (1.2\%) \\
\rowcolor{white}
 &  & phi3:latest & 27 (10.8\%) & 27 (10.8\%) & 24 (9.6\%) & 25 (10.0\%) & 28 (11.2\%) \\
\rowcolor{white}
 &  & mistral:instruct & 2 (0.8\%) & 2 (0.8\%) & 2 (0.8\%) & 2 (0.8\%) & 2 (0.8\%) \\
\rowcolor{white}
 &  & gemma:7b-instruct & --- & --- & --- & --- & --- \\
\rowcolor{white}
 &  & gpt-4 & 3 (1.2\%) & 3 (1.2\%) & 3 (1.2\%) & 3 (1.2\%) & 3 (1.2\%) \\\cmidrule{2-8}
\rowcolor{azure}  
 & Complex & llama3.1:8b & --- & --- & --- & --- & --- \\
\rowcolor{azure}
 &  & phi3:latest & --- & --- & --- & --- & --- \\
\rowcolor{azure}
 &  & mistral:instruct & --- & --- & --- & --- & --- \\
\rowcolor{azure}
 &  & gemma:7b-instruct & --- & --- & --- & --- & --- \\
\rowcolor{azure}
\midrule
\rowcolor{white}
Essays & Simple & llama3.1:8b & 5 (0.2\%) & 6 (0.24\%) & 7 (0.28\%) & 4 (0.16\%) & --- \\
\rowcolor{white}
 &  & phi3:latest & 2002 (81.15\%) & 2068 (83.83\%) & 1978 (80.18\%) & 2093 (84.84\%) & 2069 (83.87\%) \\
\rowcolor{white}
 &  & mistral:instruct & --- & --- & --- & --- & --- \\
\rowcolor{white}
 &  & gemma:7b-instruct & --- & --- & --- & --- & --- \\
\rowcolor{white}
 &  & gpt-4 & 28 (1.13\%) & 28 (1.13\%) & 28 (1.13\%) & 28 (1.13\%) & 28 (1.13\%) \\\cmidrule{2-8}
\rowcolor{azure}
 & Complex & llama3.1:8b & 2 (0.08\%) & 2 (0.08\%) & 3 (0.12\%) & 3 (0.12\%) & 1 (0.04\%) \\
\rowcolor{azure}
 &  & phi3:latest & --- & --- & --- & --- & --- \\
\rowcolor{azure}
 &  & mistral:instruct & --- & --- & --- & --- & --- \\
\rowcolor{azure}
 &  & gemma:7b-instruct & --- & --- & --- & --- & --- \\
\rowcolor{azure}
\midrule
\rowcolor{white}
Pandora & Simple & llama3.1:8b & 54 (0.38\%) & 63 (0.44\%) & 66 (0.46\%) & 44 (0.31\%) & 54 (0.38\%) \\
\rowcolor{white}
 &  & phi3:latest & 477 (3.35\%) & 348 (2.45\%) & 347 (2.44\%) & 486 (3.42\%) & 386 (2.71\%) \\
\rowcolor{white}
 &  & mistral:instruct & 3 (0.02\%) & 3 (0.02\%) & 3 (0.02\%) & 2 (0.01\%) & 3 (0.02\%) \\
\rowcolor{white}
 &  & gemma:7b-instruct & 117 (0.82\%) & 73 (0.51\%) & 96 (0.68\%) & 83 (0.58\%) & 95 (0.67\%) \\
\rowcolor{white}
 &  & gpt-4 & 679 (4.77\%) & 679 (4.77\%) & 679 (4.77\%) & 679 (4.77\%) & 679 (4.77\%) \\\cmidrule{2-8}
\rowcolor{azure}
 & Complex & llama3.1:8b & 220 (1.55\%) & 313 (2.2\%) & 294 (2.07\%) & 377 (2.65\%) & 317 (2.23\%) \\
\rowcolor{azure}
 &  & phi3:latest & 5 (0.04\%) & 3 (0.02\%) & 6 (0.04\%) & 8 (0.06\%) & 11 (0.08\%) \\
\rowcolor{azure}
 &  & mistral:instruct & 5 (0.04\%) & 4 (0.03\%) & 5 (0.04\%) & 5 (0.04\%) & 5 (0.04\%) \\
\rowcolor{azure}
 &  & gemma:7b-instruct & --- & --- & --- & --- & --- \\
\rowcolor{azure}
\bottomrule
\end{tabular}
\caption{Number (and percentage relative to dataset size) of invalid outputs, broken down by model, dataset, prompt type, and OCEAN trait.  The symbol ``---'' indicates zero invalid outputs.}
\label{tab:invalid-outputs}
\end{table*}

\begin{table*}[!h]
\centering
\small
\setlength{\tabcolsep}{3pt}
\renewcommand{\arraystretch}{0.95}
\begin{tabular}{lllccccc}
\toprule
\rowcolor{white}
\textbf{Dataset} & \textbf{Prompt} & \textbf{Model A} & \multicolumn{5}{c}{\textbf{Model B}} \\
 &  &  & llama3.1:8b & phi3:latest & mistral:instruct & gemma:7b-instruct & gpt-4 \\
\midrule
\rowcolor{white}
MyPersonality & Simple & llama3.1:8b & 3 & 1 & 2 & --- & --- \\
\rowcolor{white}
 &  & phi3:latest & 1 & 35 & 1 & --- & --- \\
\rowcolor{white}
 &  & mistral:instruct & 2 & 1 & 2 & --- & --- \\
\rowcolor{white}
 &  & gemma:7b-instruct & --- & --- & --- & --- & --- \\
\rowcolor{white}
 &  & gpt-4 & --- & --- & --- & --- & 3 \\ \cmidrule{2-8}
\rowcolor{azure}
 & Complex & llama3.1:8b & --- & --- & --- & --- & --- \\
\rowcolor{azure}
 &  & phi3:latest & --- & --- & --- & --- & --- \\
\rowcolor{azure}
 &  & mistral:instruct & --- & --- & --- & --- & --- \\
\rowcolor{azure}
 &  & gemma:7b-instruct & --- & --- & --- & --- & --- \\
\rowcolor{azure}
\midrule
\rowcolor{white}
Essays & Simple & llama3.1:8b & 8 & 7 & --- & --- & --- \\
\rowcolor{white}
 &  & phi3:latest & 7 & 2146 & --- & --- & --- \\
\rowcolor{white}
 &  & mistral:instruct & --- & --- & --- & --- & --- \\
\rowcolor{white}
 &  & gemma:7b-instruct & --- & --- & --- & --- & --- \\
\rowcolor{white}
 &  & gpt-4 & --- & --- & --- & --- & 28 \\ \cmidrule{2-8}
\rowcolor{azure}
 & Complex & llama3.1:8b & 3 & --- & --- & --- & --- \\
\rowcolor{azure}
 &  & phi3:latest & --- & --- & --- & --- & --- \\
\rowcolor{azure}
 &  & mistral:instruct & --- & --- & --- & --- & --- \\
\rowcolor{azure}
 &  & gemma:7b-instruct & --- & --- & --- & --- & --- \\
\rowcolor{azure}
\midrule
\rowcolor{white}
Pandora & Simple & llama3.1:8b & 85 & 17 & 2 & 10 & 33 \\
\rowcolor{white}
 &  & phi3:latest & 17 & 691 & 2 & 32 & 29 \\
\rowcolor{white}
 &  & mistral:instruct & 2 & 2 & 3 & 2 & --- \\
\rowcolor{white}
 &  & gemma:7b-instruct & 10 & 32 & 2 & 123 & 5 \\
\rowcolor{white}
 &  & gpt-4 & 33 & 29 & --- & 5 & 480 \\ \cmidrule{2-8}
\rowcolor{azure}
 & Complex & llama3.1:8b & 445 & 7 & 5 & --- & --- \\
\rowcolor{azure}
 &  & phi3:latest & 7 & 18 & 1 & --- & --- \\
\rowcolor{azure}
 &  & mistral:instruct & 5 & 1 & 7 & --- & --- \\
\rowcolor{azure}
 &  & gemma:7b-instruct & --- & --- & --- & --- & --- \\
\rowcolor{azure}
\bottomrule
\end{tabular}
\caption{Shared invalid outputs (raw counts) between models by dataset and prompt. The symbol ``---'' indicates zero shared invalid outputs.}
\label{tab:shared-invalid-outputs}
\end{table*}


\vspace{-2pt}
\section{Results}
\label{sec:results}

We report the results of 135 experiments, spanning two prompting strategies (for all models but GPT-4), three datasets, five traits, and five LLMs. Analyses proceed in two stages: first, by examining invalid outputs (failures and co-failures), and then by evaluating valid predictions using both aggregate and class-wise metrics.

\vspace{-2pt}
\subsection{Invalid Outputs}

Before assessing predictive performance, we evaluate the reliability of model outputs across prompt types, datasets, and traits. An output is considered valid if it consists solely of, or begins with, either 0 or 1.

Table~\ref{tab:invalid-outputs} summarizes the invalid outputs across the experimental setups, that is, the number of texts for which the LLM produced an invalid response. For each dataset, prompt type, and model, the table reports the number of invalid outputs per trait, with percentages calculated relative to the dataset size.
Under the simple prompt, most LLMs produce at least some invalid outputs, with Phi-3 yielding a particularly large number on the Essays dataset. Failures appear dataset-specific rather than trait-specific: within a dataset, counts are nearly identical across traits but vary substantially between datasets. The main outlier is Phi-3, which shows moderate invalid rates in MyPersonality ($\sim$11.2\%) and Pandora ($\sim$3\%), but as high as $\sim$80\% in Essays. Other models are generally reliable ($\sim$1\%), with slightly elevated rates in Pandora. Gemma (errors only in Pandora), Mistral (no errors in Essays), and LLaMA 3 are the most robust overall. GPT-4 produces fewer invalid outputs than Phi-3, but remains sensitive in Pandora. Note that although the absolute number of invalid outputs is higher for Pandora—being by far the largest dataset—the proportions (i.e., percentages) are fairly similar across datasets.

Complex prompts reduce invalid outputs dramatically, often eliminating them entirely. Phi-3, previously the least reliable, becomes fully robust in Essays and nearly so in Pandora. However, this improvement is not universal: LLaMA 3 and Mistral show slight regressions in Pandora, with LLaMA~3 generating more than 200 additional invalid predictions compared to the simple prompt. An unusual case arises in Essays for Neuroticism, where one invalid output emerges under the complex prompt that was absent before.

\begin{tcolorbox}[keyfindingbox, title=Key Finding 1: Enriched prompts reduce invalid outputs.]
Complex prompts greatly improve output validity across models and datasets, reducing invalid responses to near zero. Phi-3, which failed on $\sim$80\% of Essays inputs, becomes fully reliable. However, exceptions exist: LLaMA 3 in Pandora produces more invalid outputs under the complex prompt, showing that richer instructions do not always guarantee improved task adherence.
\end{tcolorbox}

\subsection{Shared Invalid Outputs}

Table~\ref{tab:shared-invalid-outputs} analyzes the overlap of invalid outputs across models. For each dataset and prompt type, it compares pairs of models and counts one for each text where both models fail on at least one trait. As an example, consider three texts and two models: if the first model fails on all traits for the first text (\texttt{100}), and the second model fails on trait O for the first text, trait C for the second, and trait E for the third (\texttt{111}), their AND is \texttt{100}, corresponding to one shared failure --- that is, the failure is shared in terms of unique texts, regardless of the traits. Here, sequences such as \texttt{100} denote binary (logical) vectors, with each position representing a text. On the diagonal (i.e., when a model is compared with itself), this reduces to counting the number of texts on which at least one failure occurs, due to the idempotency of the logical AND.
The table also reports the dataset, prompt type, and compared models. Diagonal values are close to those in Table~\ref{tab:invalid-outputs}, confirming that most errors occur on the same texts across traits. For example, LLaMA 3 in MyPersonality consistently fails on the same three texts regardless of trait, whereas Phi-3 shows only partial overlap across traits, resulting in a larger number of unique failures.

Table~\ref{tab:shared-invalid-outputs} also reveals different failure modes. Some models, such as Mistral and Phi3 in Pandora under the complex prompt, fail only on inputs that also trigger errors in other models. Others, like GPT-4 and Phi-3 in Pandora under the simple prompt, fail independently and on distinct inputs. In MyPersonality, GPT-4 produces three unique failures, while Mistral’s two errors are fully shared with LLaMA 3. This pattern indicates that invalid outputs can arise from different underlying mechanisms. By manually inspecting a subsample of failures, we noticed that invalid outputs are driven by safety filters in GPT-4, while Phi-3 shows a mix of incoherent completion (e.g., task misunderstanding, verbose answer) and refusals (e.g., lack of context).

\begin{tcolorbox}[keyfindingbox, title=Key Finding 2: Failure modes differ across models.]
Some models (e.g., Mistral and Phi3 in Pandora under the complex prompt) produce invalid outputs only on inputs problematic for other LLMs, while others (e.g., GPT-4 and Phi-3 in Pandora under the simple prompt) fail independently on unique inputs. These divergent patterns suggest that invalid outputs stem from model-specific vulnerabilities, not just shared difficulties in the data.
\end{tcolorbox}

\subsection{Valid Predictions}

\begin{figure}[htpb]
    \centering
    \includegraphics[width=0.4\linewidth]{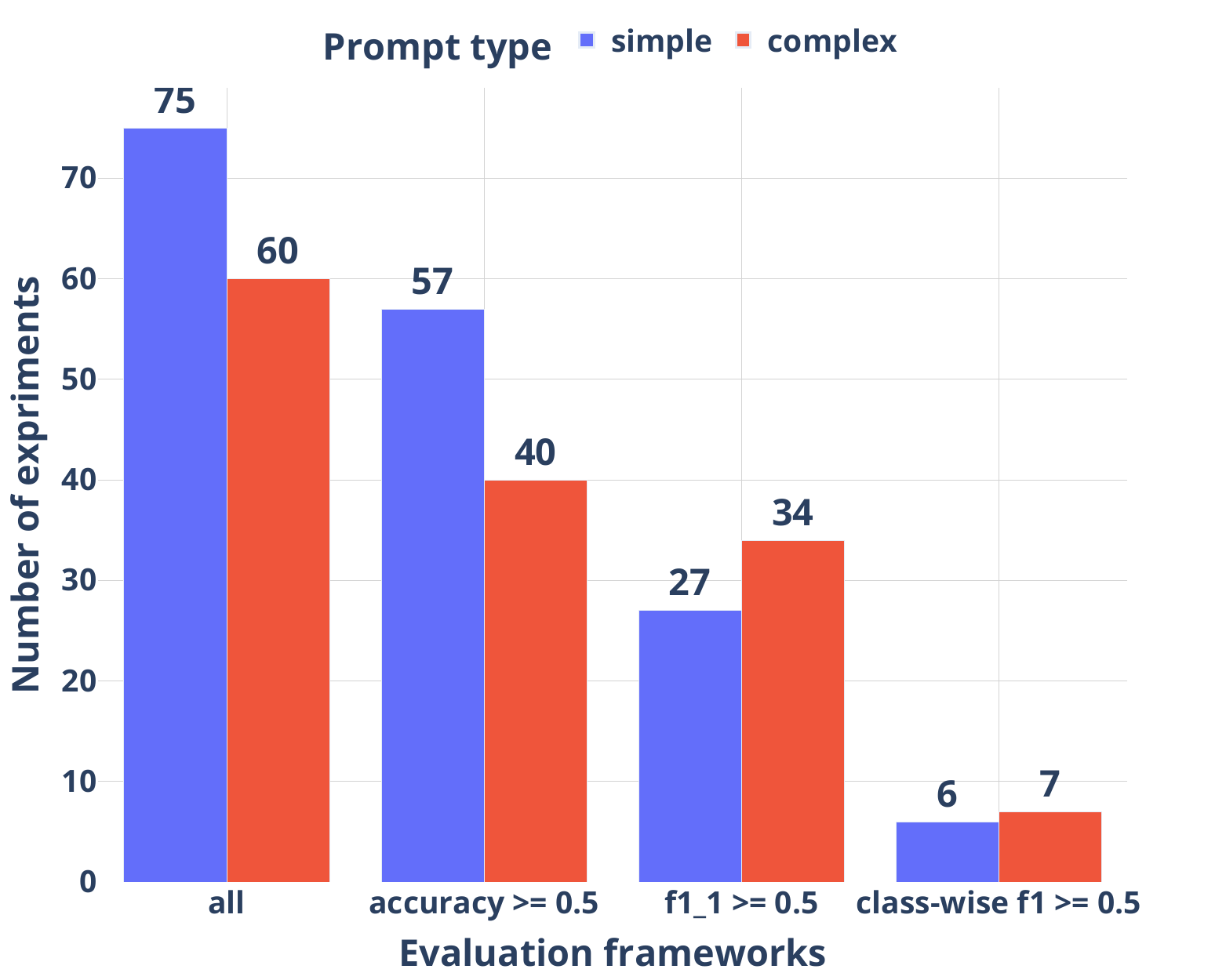}
    \caption{Number of experiments per prompt strategy and evaluation framework}
    \label{fig:num-experiments}
\end{figure}

We conducted 135 experiments in total: 75 with the simple prompt and 60 with the complex one\footnote{The number of experiments with simple and complex prompts varies, as we do not try the complex prompt strategy with GPT4.} (Figure~\ref{fig:num-experiments}). In each experiment, invalid outputs were discarded before evaluation, so all reported results are based exclusively on valid predictions.
To identify configurations with at least minimally informative predictions, we apply a conservative threshold of 0.5 on both accuracy and F1. While this threshold does not represent true chance performance under class imbalance, it serves as a safeguard against trivial solutions (e.g., predicting only the majority class). Applying this filter progressively reduces the number of informative experiments: first when considering accuracy, then F1, and finally class-wise F1. Ultimately, only 13 experiments (6 with the simple prompt, 7 with the complex) meet the class-wise F1 threshold ($\geq$0.5 on both classes), suggesting at least weak discrimination between presence (1) and absence (0) of traits.

\subsubsection{Class-wise F1}

\begin{figure*}[!htbp]
  \centering
  \begin{minipage}[t]{0.49\textwidth}
    \centering
    \includegraphics[width=\linewidth]{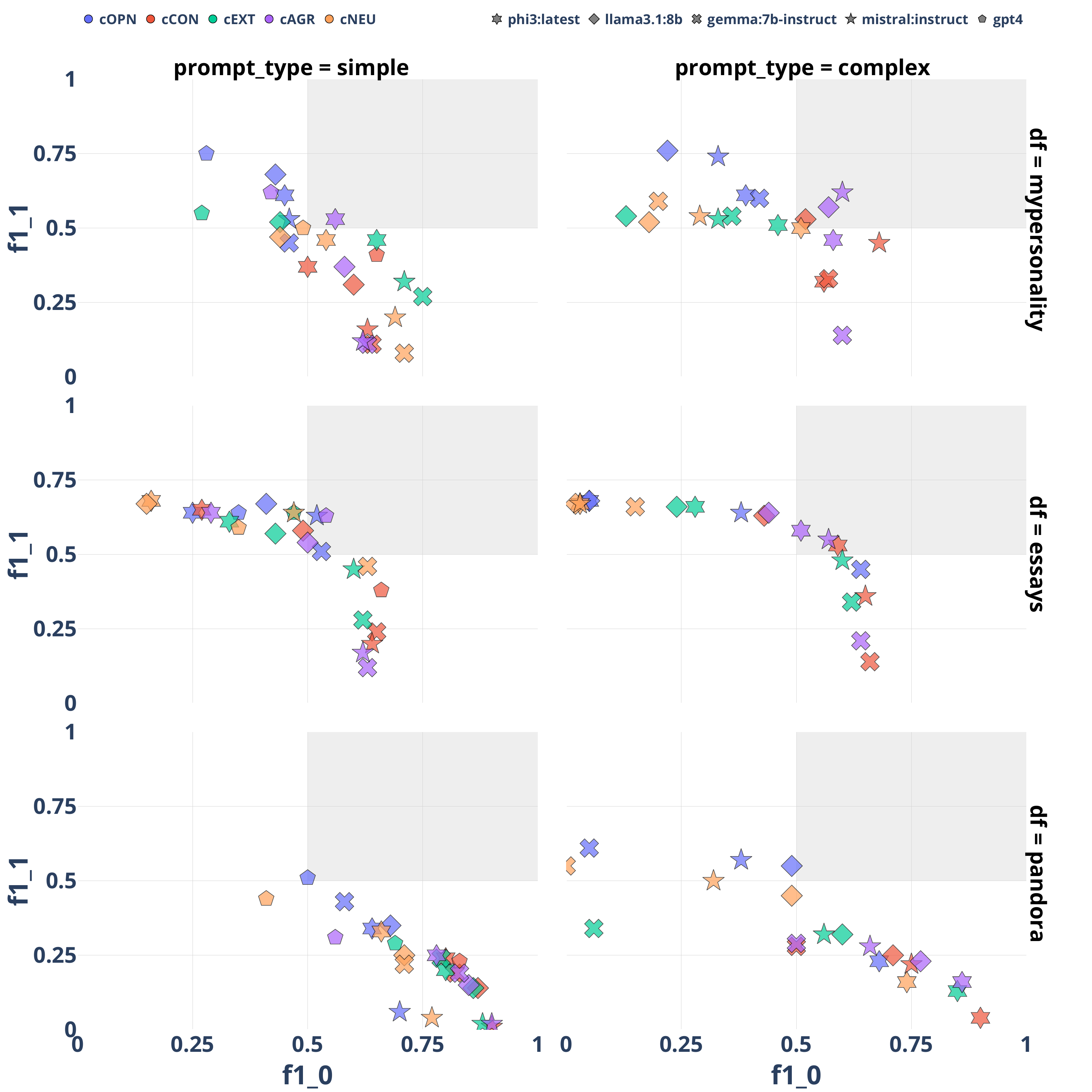}
    \caption{Evaluation of LLMs according to class-wise F1.}
    \label{fig:scatter-f1}
  \end{minipage}
  \hfill
  \begin{minipage}[t]{0.49\textwidth}
    \centering
    \includegraphics[width=\linewidth]{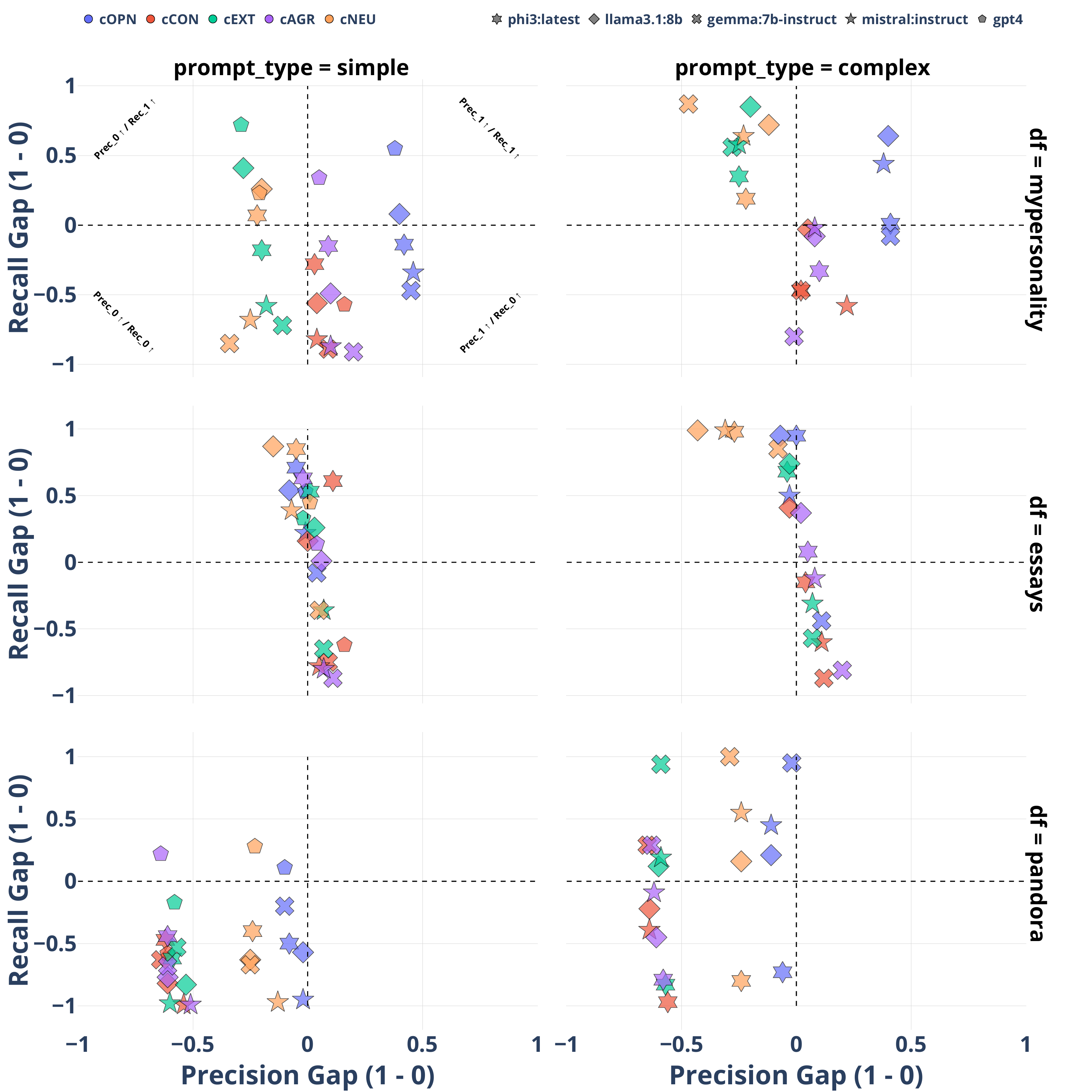}
    \caption{Evaluation of LLMs according to precision and recall gaps.}
    \label{fig:scatter-prec-recall}
  \end{minipage}

\end{figure*}

Figure~\ref{fig:scatter-f1} shows class-wise F1 across models, traits, datasets, and prompting strategies. Dashed lines mark the $\geq 0.5$ threshold. Only 13 experiments surpass it, mostly involving Agreeableness and Openness in Essays and MyPersonality. Complex prompts rarely boost already strong models, but they often help weaker ones recover balance. For instance, Mistral predicting Agreeableness in MyPersonality improves from a heavily biased profile (low $F1_1$, high $F1_0$) under the simple prompt to the most balanced result across all experiments ($\sim$0.6 for both classes). A similar effect appears in Essays.

Another consistent pattern across datasets in our experiments is that not all traits are equally difficult to detect. Openness and Agreeableness emerge as relatively easier to capture, with some experiments approaching balanced F1 values, while Extraversion and Neuroticism remain the most elusive, rarely exceeding the 0.5 threshold even under enriched prompts. Conscientiousness typically falls in between, showing occasional improvements but without stable gains across models and datasets. It is a common finding in previous work that some traits are easier to predict than others. However, the specific traits that are easier to predict vary across datasets and depend on the APPT methodology applied.

\begin{tcolorbox}[keyfindingbox, title=Key Finding 3: Complex prompts help weaker models balance predictions.]
Enriched prompts do not consistently improve high-performing models but can enable weaker ones to recover balanced detection. Mistral predicting Agreeableness in MyPersonality is the best case, with class-wise F1 $\sim$0.6 for both classes, highlighting the corrective effect of enriched linguistic cues.
\end{tcolorbox}

Interestingly, in some cases open-source lightweight models outperform GPT-4 under the simple prompt (e.g., Phi-3 on Neuroticism in MyPersonality, LLaMA 3 on Openness in Pandora). With complex prompts, these models approximate GPT-4’s best results, suggesting that prompt design can close the gap.

\begin{tcolorbox}[keyfindingbox, title=Key Finding 4: Prompt enrichment allows open-source models to match GPT-4 in its strongest cases.]
In traits where GPT-4 performs best (Neuroticism in MyPersonality, Agreeableness in Essays, Openness in Pandora), open-source models such as Phi-3 and LLaMA 3 approach or match its performance under complex prompts. Prompt design thus acts as a powerful equalizer.
\end{tcolorbox}

\subsubsection{Precision and Recall Gaps}

Figure~\ref{fig:scatter-prec-recall} shows precision and recall gaps across traits and datasets. Precision gaps largely mirror label distributions: balanced traits yield near-zero precision gaps (e.g., all traits in Essays), while skewed traits induce negative or positive gaps consistent with majority-class bias. Recall gaps, however, are more variable. Even when precision is balanced, recall shows large divergences, revealing that models cannot consistently distinguish between classes beyond chance.

\begin{tcolorbox}[keyfindingbox, title=Key Finding 5: Poor performance is driven by recall\, not precision.]
Precision gaps mostly reflect dataset distributions, but recall behavior varies widely across models and traits. This indicates that low F1 scores stem from models’ inability to recall minority-class instances, rather than from precision asymmetries.
\end{tcolorbox}

\subsubsection{Best Results}

Closer inspection of the 13 experiments with class-wise F1 $\geq 0.5$ highlights the rarity of balanced predictions. Mistral on Agreeableness in MyPersonality under the complex prompt is the only case with consistently reliable performance across metrics. By contrast, GPT-4 on Agreeableness in Essays achieves similar accuracy but reveals severe recall imbalance. Likewise, GPT-4 on Openness in Pandora suffers from recall asymmetry despite a balanced distribution.  

Overall, apparent aggregate successes often mask bias, with Mistral’s Agreeableness result standing out as the only genuinely balanced case.

\subsubsection{Prompt Effects}

\begin{figure*}[!p]
  \centering
  \begin{subfigure}[t]{0.48\textwidth}
    \centering
    \includegraphics[width=\linewidth]{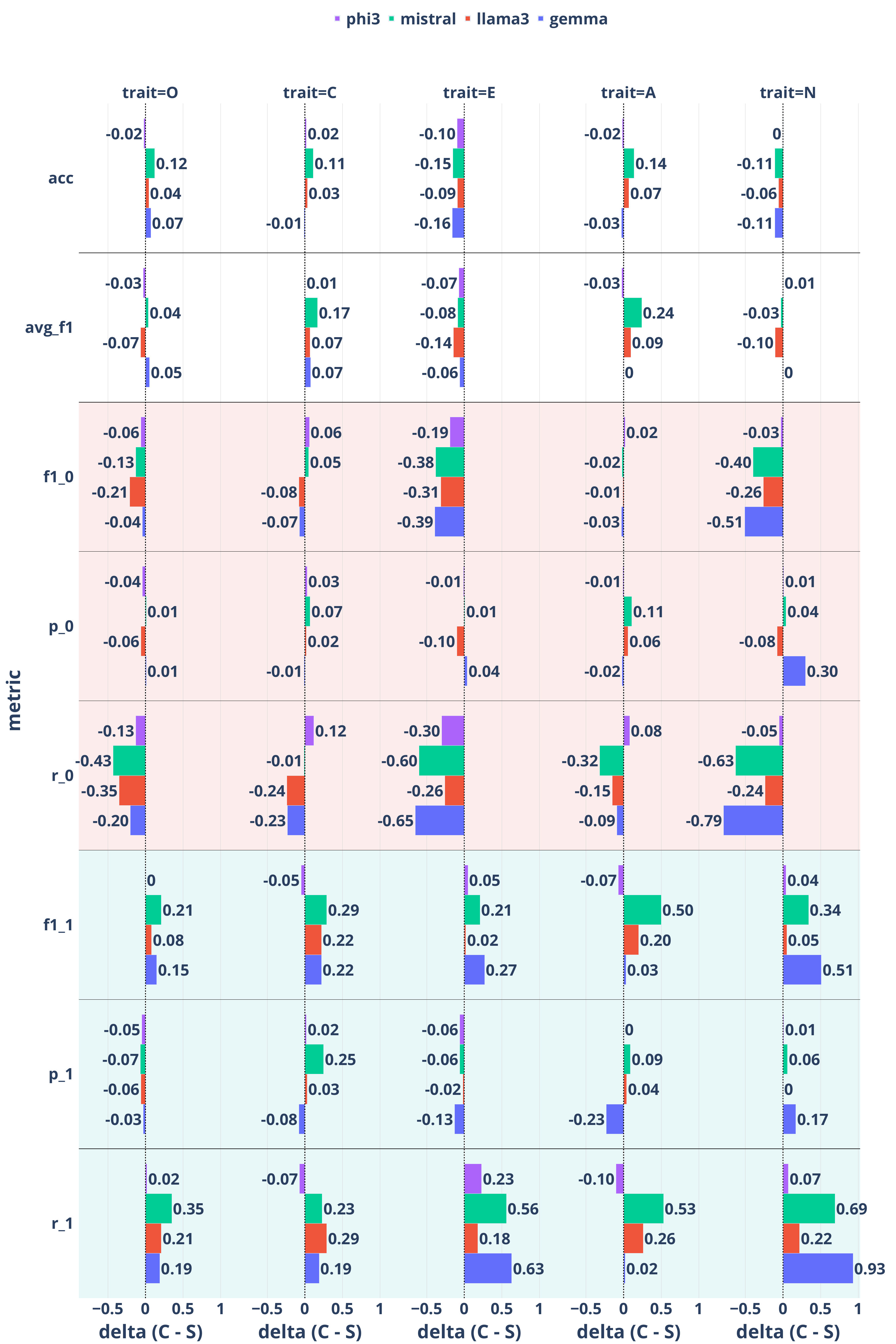}
    \caption{Mypersonality}
    \label{fig:delta_mypers}
  \end{subfigure}
  \hfill
  \begin{subfigure}[t]{0.48\textwidth}
    \centering
    \includegraphics[width=\linewidth]{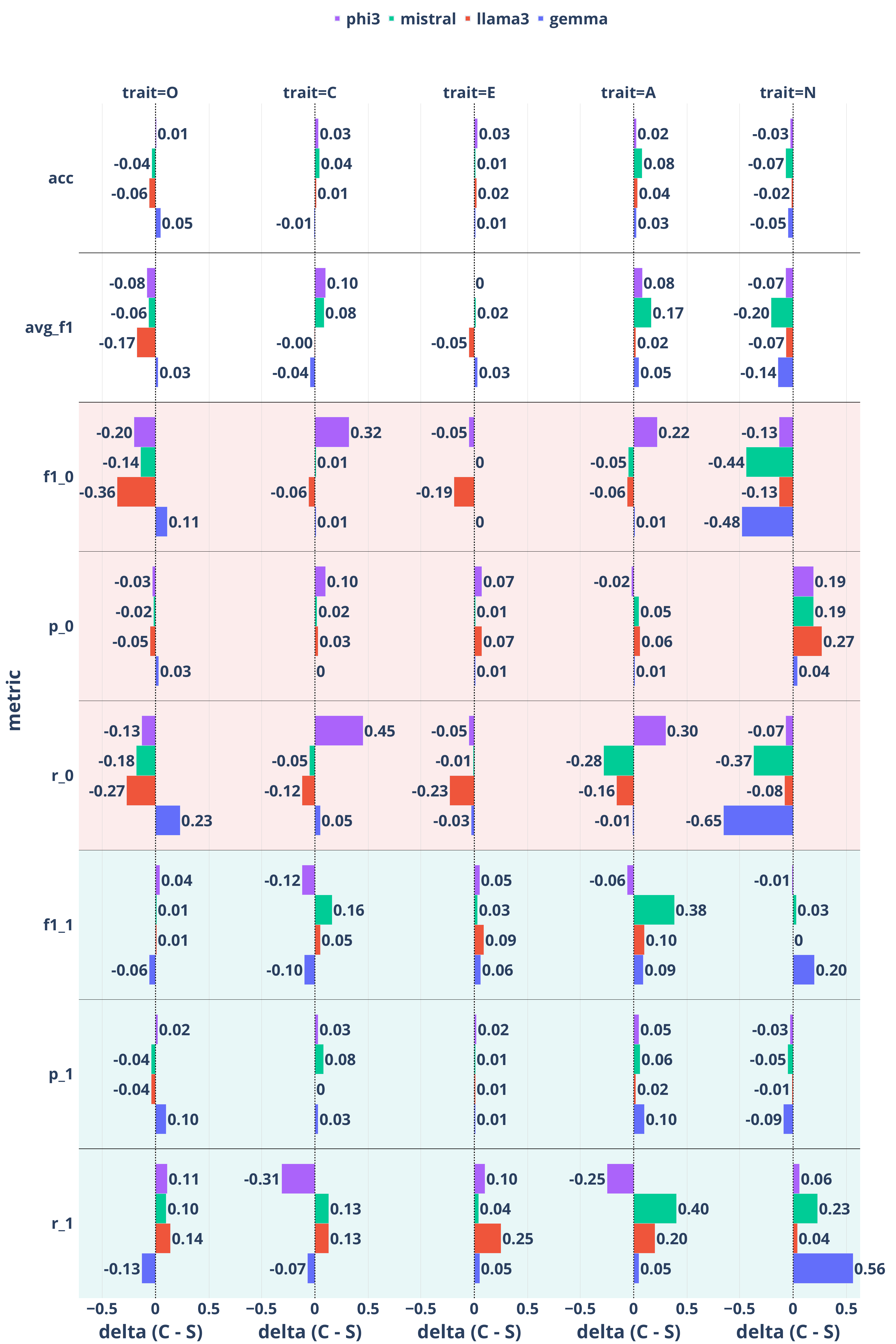}
    \caption{Essays}
    \label{fig:delta_essays}
  \end{subfigure}
  \caption{Differences in the metrics of full binary classification report (rows), per trait (columns) and models (bars). The x values indicate the gap between the results of complex and simple prompt (i.e., Complex - Simple) w.r.t. each evaluation metric. Note that the plot relative to Pandora has been moved to Appendix in Figure~\ref{fig:pandora_delta} due to space constraints }
  \label{fig:prompt_delta}
\end{figure*}

Finally, Figure~\ref{fig:prompt_delta} compares complex versus simple prompts. Gains in accuracy are limited and inconsistent. More revealing are shifts in class-wise F1: complex prompts almost always improve the positive class (Recall\_1, F1\_1) while degrading the negative class (Recall\_0, F1\_0). This systematic positive bias is evident across datasets, though in some cases it counterbalances an existing negative bias, leading to more balanced predictions (again, Mistral on Agreeableness in MyPersonality is the clearest example).

\begin{tcolorbox}[keyfindingbox, title=Key Finding 6: Complex prompts induce positive bias.]
Complex prompts systematically increase recall for the positive class, biasing predictions toward trait presence. While this often skews results, in some cases it offsets a prior negative bias, producing more balanced outcomes.
\end{tcolorbox}

\section{Conclusion}
\label{sec:conclusion}

We presented a comprehensive evaluation of LLMs for binary APPT under the Big Five, spanning five models, three datasets, and two prompting regimes (minimal vs.\ enriched). Our results show that \emph{prompt design is pivotal}—not only for output validity, but also for the balance between detecting trait presence and absence. Enriched prompts nearly eliminate invalid outputs and can help underperforming models recover recall on the positive class; however, this benefit often introduces a \emph{systematic positive bias}. In select cases, that shift is beneficial: it counteracts a prior negative skew and yields the most balanced outcomes (e.g., Mistral on Agreeableness in MyPersonality).
We further find that lightweight open-source models can \emph{approach} GPT-4 in their strongest configurations when provided with enriched prompts, yet no model consistently surpasses pre-LLM benchmarks in our zero-shot setting. Thus, while LLMs are promising tools, they do not constitute a paradigm shift for binary APPT as currently framed.
Methodologically, our analyses underscore that aggregate scores (accuracy, macro-F1) can mask substantial asymmetries. \emph{Per-class recall} emerges as the most informative indicator of real progress, revealing where models truly succeed or fail across classes. Performance variability across traits and datasets also points to a strong dependence on how (and whether) personality is linguistically expressed in different contexts.

\paragraph{Limitations and recommendations.}
Our study is limited to zero-shot prompting and a binary framing, which simplify evaluation but compress nuanced psychological constructs and interact with class imbalance. Nonetheless, these constraints have value: the binary setup is widely used in APPT, allowing for direct comparison with pre-LLM baselines, while the zero-shot condition reflects realistic scenarios where labeled data are scarce and models are used “out of the box".  

Another limitation is that enriched prompts, while reducing invalid outputs, induce a systematic positive bias toward predicting trait presence. This, however, is informative in itself: it highlights how prompt design can shape model behavior and underscores the need for calibration and evaluation choices that account for such biases.  

We also excluded very large models (e.g., Grok, Falcon) due to resource constraints, and the GPT-4 setup differed slightly from that of open-source LLMs --- but only in terms of hard rules to increase the validity of open-source outputs. Yet the focus on lightweight open-source models is pragmatic, as they represent resource-efficient alternatives, and the GPT-4 comparison still reveals where smaller systems can approach proprietary performance.  


Building on these observations, we recommend that future work: (i) report trait and text distributions alongside \emph{class-wise} metrics; (ii) disentangle validity, bias, and performance (e.g., by auditing refusals and invalid outputs); (iii) explore calibrated decision rules, cost-sensitive objectives, and lightweight adaptation (e.g., few-shot, PEFT) to mitigate bias–recall trade-offs.

Overall, effective APPT with LLMs will require \emph{coordinating prompt design, metric choice, and trait-specific analysis}. Without such coordination, gains in validity or headline scores risk coming at the expense of balanced, interpretable predictions.

\bibliographystyle{plainnat}
\bibliography{bibliography.bib}

@article{costa1992normal,
  title={Normal personality assessment in clinical practice: The NEO Personality Inventory.},
  author={Costa, Paul T and McCrae, Robert R},
  journal={Psychological assessment},
  volume={4},
  number={1},
  pages={5},
  year={1992},
  publisher={American Psychological Association}
}

@book{galton1884measurement,
  author    = {Francis Galton},
  title     = {Measurement of Character},
  year      = {1884},
  journal   = {Fortnightly Review},
  volume    = {36},
  pages     = {179-185}
}

@article{allport1936trait,
  author    = {Gordon W. Allport and Henry S. Odbert},
  title     = {Trait-names: A psycho-lexical study},
  journal   = {Psychological Monographs},
  volume    = {47},
  number    = {1},
  pages     = {1-171},
  year      = {1936},
  publisher = {American Psychological Association},
  doi       = {10.1037/h0093360}
}

@book{brunswik1955perception,
  author    = {Egon Brunswik},
  title     = {Perception and the Representative Design of Psychological Experiments},
  year      = {1955},
  publisher = {University of California Press},
  address   = {Berkeley, CA}
}

@article{kosinski2013private,
  title={Private traits and attributes are predictable from digital records of human behavior},
  author={Kosinski, Michal and Stillwell, David and Graepel, Thore},
  journal={Proceedings of the national academy of sciences},
  volume={110},
  number={15},
  pages={5802--5805},
  year={2013},
  publisher={National Academy of Sciences}
}

@article{gjurkovic2020pandora,
  title={PANDORA talks: Personality and demographics on Reddit},
  author={Gjurkovi{\'c}, Matej and Karan, Mladen and Vukojevi{\'c}, Iva and Bo{\v{s}}njak, Mihaela and {\v{S}}najder, Jan},
  journal={arXiv preprint arXiv:2004.04460},
  year={2020}
}

@incollection{goldberg2013alternative,
  title={An alternative “description of personality”: The Big-Five factor structure},
  author={Goldberg, Lewis R},
  booktitle={Personality and personality disorders},
  pages={34--47},
  year={2013},
  publisher={Routledge}
}

@article{peters2024large,
  title={Large language models can infer personality from free-form user interactions},
  author={Peters, Heinrich and Cerf, Moran and Matz, Sandra C},
  journal={arXiv preprint arXiv:2405.13052},
  year={2024}
}

@article{piastra2025emergent,
  title={On the emergent capabilities of ChatGPT 4 to estimate personality traits},
  author={Piastra, Marco and Catellani, Patrizia},
  journal={Frontiers in Artificial Intelligence},
  volume={8},
  pages={1484260},
  year={2025},
  publisher={Frontiers Media SA}
}

@article{zhu2025investigating,
  title={Investigating Large Language Models in Inferring Personality Traits from User Conversations},
  author={Zhu, Jianfeng and Jin, Ruoming and Coifman, Karin G},
  journal={arXiv preprint arXiv:2501.07532},
  year={2025}
}

@article{yan2024predicting,
  title={Predicting the big five personality traits in chinese counselling dialogues using large language models},
  author={Yan, Yang and Ma, Lizhi and Li, Anqi and Ma, Jingsong and Lan, Zhenzhong},
  journal={arXiv preprint arXiv:2406.17287},
  year={2024}
}

@article{ji2023chatgpt,
  title={Is chatgpt a good personality recognizer? a preliminary study},
  author={Ji, Yu and Wu, Wen and Zheng, Hong and Hu, Yi and Chen, Xi and He, Liang},
  journal={arXiv preprint arXiv:2307.03952},
  year={2023}
}

@article{derner2024can,
  title={Can ChatGPT read who you are?},
  author={Derner, Erik and Ku{\v{c}}era, Dalibor and Oliver, Nuria and Zah{\'a}lka, Jan},
  journal={Computers in Human Behavior: Artificial Humans},
  volume={2},
  number={2},
  pages={100088},
  year={2024},
  publisher={Elsevier}
}

@article{amin2023will,
  title={Will affective computing emerge from foundation models and general artificial intelligence? A first evaluation of ChatGPT},
  author={Amin, Mostafa M and Cambria, Erik and Schuller, Bj{\"o}rn W},
  journal={IEEE Intelligent Systems},
  volume={38},
  number={2},
  pages={15--23},
  year={2023},
  publisher={IEEE}
}

@article{ganesan2023systematic,
  title={Systematic evaluation of GPT-3 for zero-shot personality estimation},
  author={Ganesan, Adithya V and Lal, Yash Kumar and Nilsson, August H{\aa}kan and Schwartz, H Andrew},
  journal={arXiv preprint arXiv:2306.01183},
  year={2023}
}

@article{pennebaker1999linguistic,
  title={Linguistic styles: language use as an individual difference.},
  author={Pennebaker, James W and King, Laura A},
  journal={Journal of personality and social psychology},
  volume={77},
  number={6},
  pages={1296},
  year={1999},
  publisher={American Psychological Association}
}

@inproceedings{rangel2015overview,
  title={Overview of the 3rd Author Profiling Task at PAN 2015},
  author={Rangel Pardo, Francisco Manuel and Celli, Fabio and Rosso, Paolo and Potthast, Martin and Stein, Benno and Daelemans, Walter},
  booktitle={CLEF 2015 evaluation labs and workshop working notes papers},
  pages={1--8},
  year={2015}
}

@article{goldberg1992development,
  title={The development of markers for the Big-Five factor structure.},
  author={Goldberg, Lewis R},
  journal={Psychological assessment},
  volume={4},
  number={1},
  pages={26},
  year={1992},
  publisher={American Psychological Association}
}

@book{matthews2003personality,
  title={Personality traits},
  author={Matthews, Gerald and Deary, Ian J and Whiteman, Martha C},
  year={2003},
  publisher={Cambridge University Press}
}

@article{zhu2022lexical,
  title={A lexical psycholinguistic knowledge-guided graph neural network for interpretable personality detection},
  author={Zhu, Yangfu and Hu, Linmei and Ning, Nianwen and Zhang, Wei and Wu, Bin},
  journal={Knowledge-Based Systems},
  volume={249},
  pages={108952},
  year={2022},
  publisher={Elsevier}
}

@article{park2015automatic,
  title={Automatic personality assessment through social media language.},
  author={Park, Gregory and Schwartz, H Andrew and Eichstaedt, Johannes C and Kern, Margaret L and Kosinski, Michal and Stillwell, David J and Ungar, Lyle H and Seligman, Martin EP},
  journal={Journal of personality and social psychology},
  volume={108},
  number={6},
  pages={934},
  year={2015},
  publisher={American Psychological Association}
}

@article{tandera2017personality,
  title={Personality prediction system from facebook users},
  author={Tandera, Tommy and Suhartono, Derwin and Wongso, Rini and Prasetio, Yen Lina and others},
  journal={Procedia computer science},
  volume={116},
  pages={604--611},
  year={2017},
  publisher={Elsevier}
}

@article{ramezani2022text,
  title={Text-based automatic personality prediction using KGrAt-Net: a knowledge graph attention network classifier},
  author={Ramezani, Majid and Feizi-Derakhshi, Mohammad-Reza and Balafar, Mohammad-Ali},
  journal={Scientific reports},
  volume={12},
  number={1},
  pages={21453},
  year={2022},
  publisher={Nature Publishing Group UK London}
}

@online{Strickland2024WhatIsGenerativeAI,
  author       = {Eliza Strickland},
  title        = {What Is Generative AI?},
  year         = {2024},
  month        = {February 14},
  url          = {https://spectrum.ieee.org/what-is-generative-ai},
  organization = {IEEE Spectrum},
  note         = {Accessed: 2025-09-11}
}

\clearpage

\appendix


 \section{Full binary classification report for the best experiments}

 In order to have the clearest interpretation of our result, we provide the full binary classification report for both the experiments matching or exceeding the 0.5 significance threshold on class-wise F1 (Table~\ref{tab:good-table}). We also provide details for the experiments matching or exceeding the significance threshold on the most commonly used set of metrics --- that is, Accuracy and $F1_1$ (Table~\ref{tab:misleading-table}). This allows us to understand which models are actually good in discriminating between classes, and which are those that would have been considered good if ignoring class-wise evaluation.

 \section{Pandora's performance delta with complex prompt}
While Essays and MyPersonality are retained in the main text as they are two very different datasets (stream-of-consciousness and short showcase platform posts), Pandora has been moved to the Appendix in Figure \ref{fig:pandora_delta} due to space constraints, as it is a hybrid between longer and articulate texts, and short online posts.

\section{Visual comparison of full binary classification reports between simple and complex prompts}
\begin{figure*}[h]
    \centering
    \includegraphics[width=0.7\linewidth]{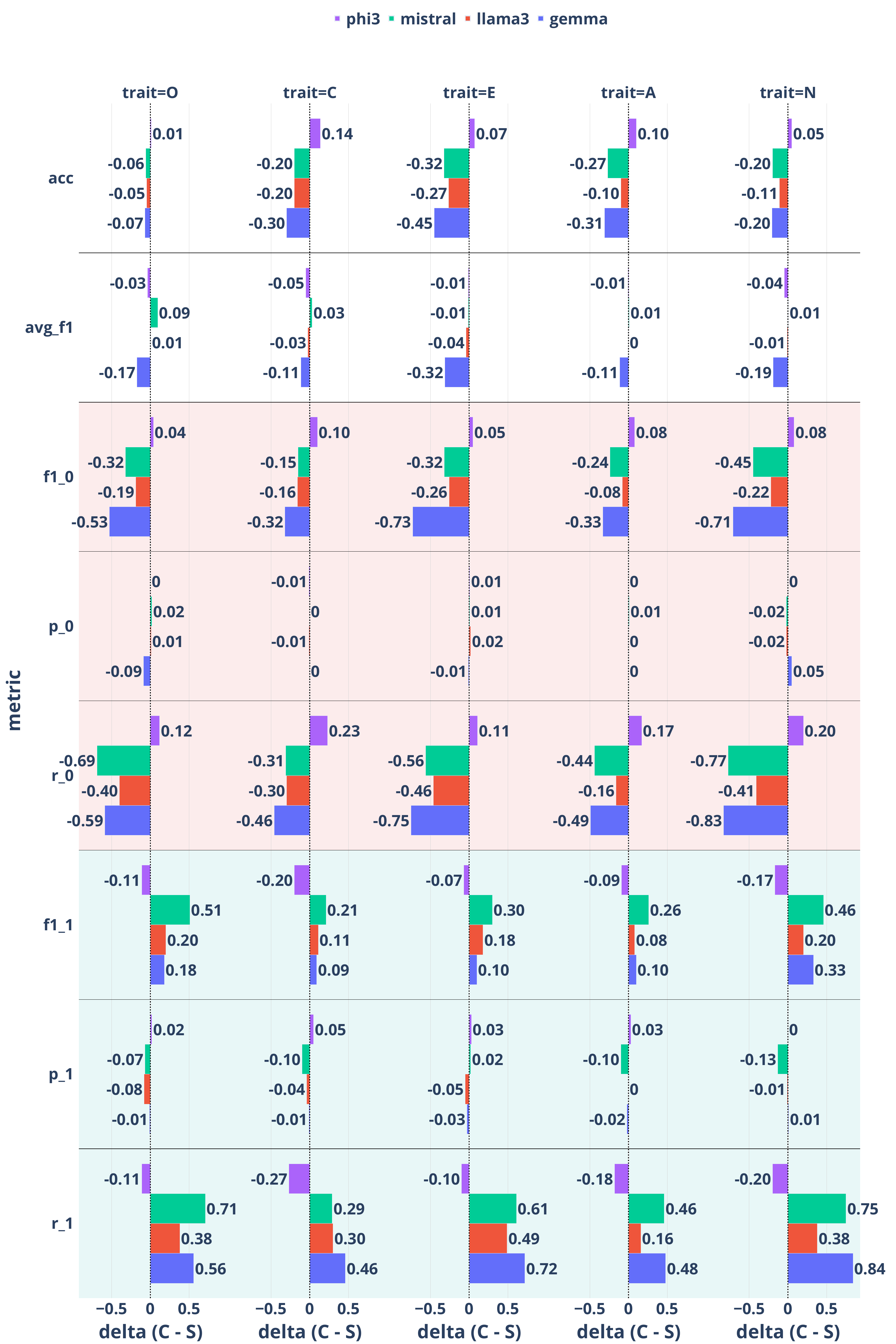}
    \caption{\textbf{Pandora only} --- Differences in the metrics of full binary classification report (rows), per trait (columns) and models (bars). The x values
indicate the gap between the results of complex and simple prompt (i.e., Complex - Simple) w.r.t. each evaluation metric.}
    \label{fig:pandora_delta}
  \end{figure*}

\begin{figure*}
    \centering
    \includegraphics[width=0.7\linewidth]{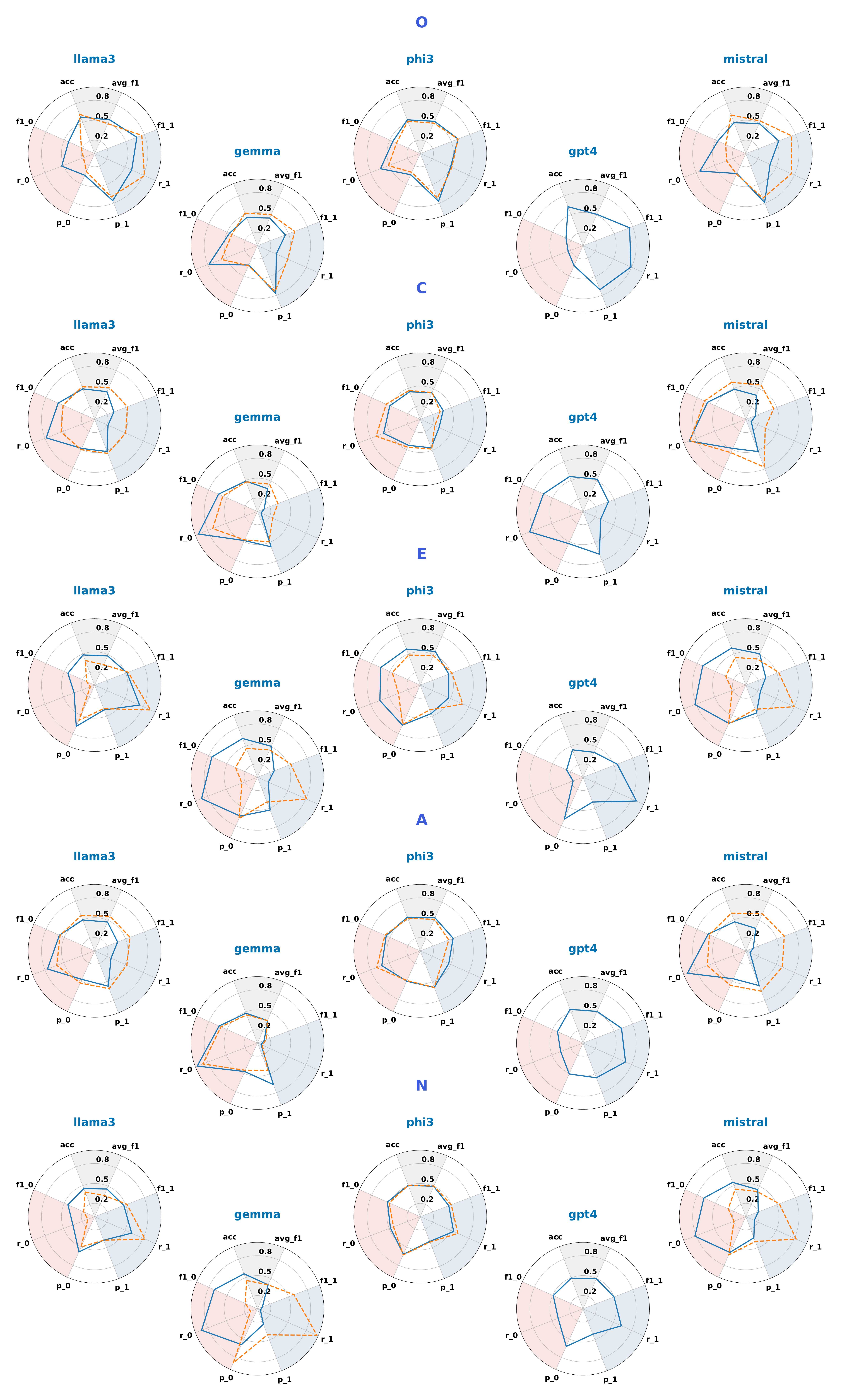}
    \caption{\textbf{MyPersonality} --- Comparison of classification reports between simple (blue solid line) and complex (orange dashed line) prompting strategies. Metrics are arrenged so that: the top part shows macro metrics (accuracy and average F1), the left and the right sides the negative and positive metrics (F1\_0, recall\_0 and precision\_0). Mistral predicting Agreeableness in Mypersonality is an example of significant improvement in both terms of prediction quality and balanced metrics. In other words, from a left-skewed shape, the complex prompt draws a hexagon close to 0.6\%. The same balancing of metrics happens also in other cases (such as Mistral predicting Agreeableness in Essays and Llama3 predicting Openness, see Figure \ref{fig:prompt_radars_2}) but with poorer values ($\sim$0.5).}
    \label{fig:prompt_radars_1}
\end{figure*}

\begin{figure*}
    \centering
    \includegraphics[width=0.7\linewidth]{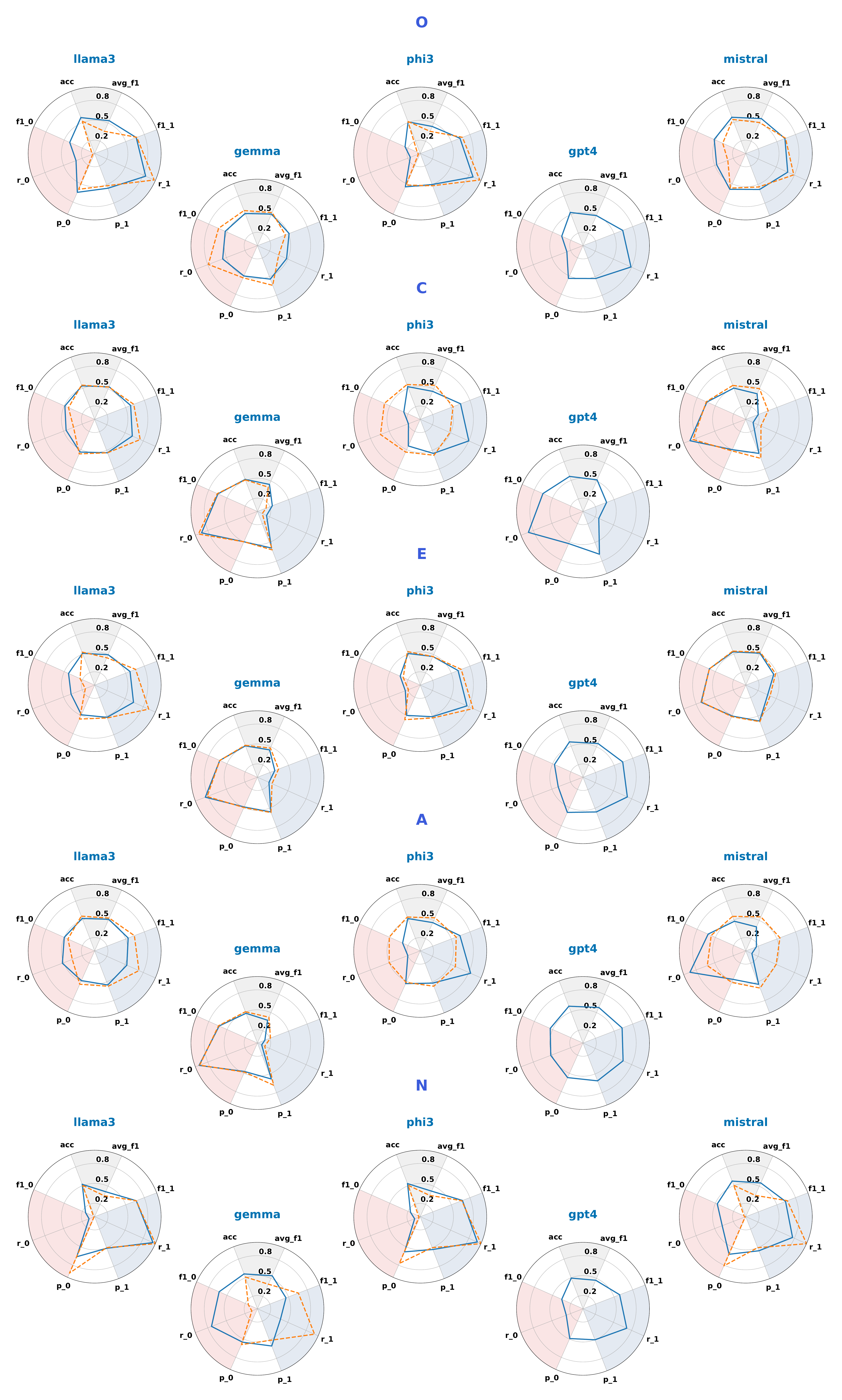}
    \caption{\textbf{Essays} --- Description in the caption of Figure \ref{fig:prompt_radars_1}}
    \label{fig:prompt_radars_2}
\end{figure*}

\begin{figure*}
    \centering
    \includegraphics[width=0.7\linewidth]{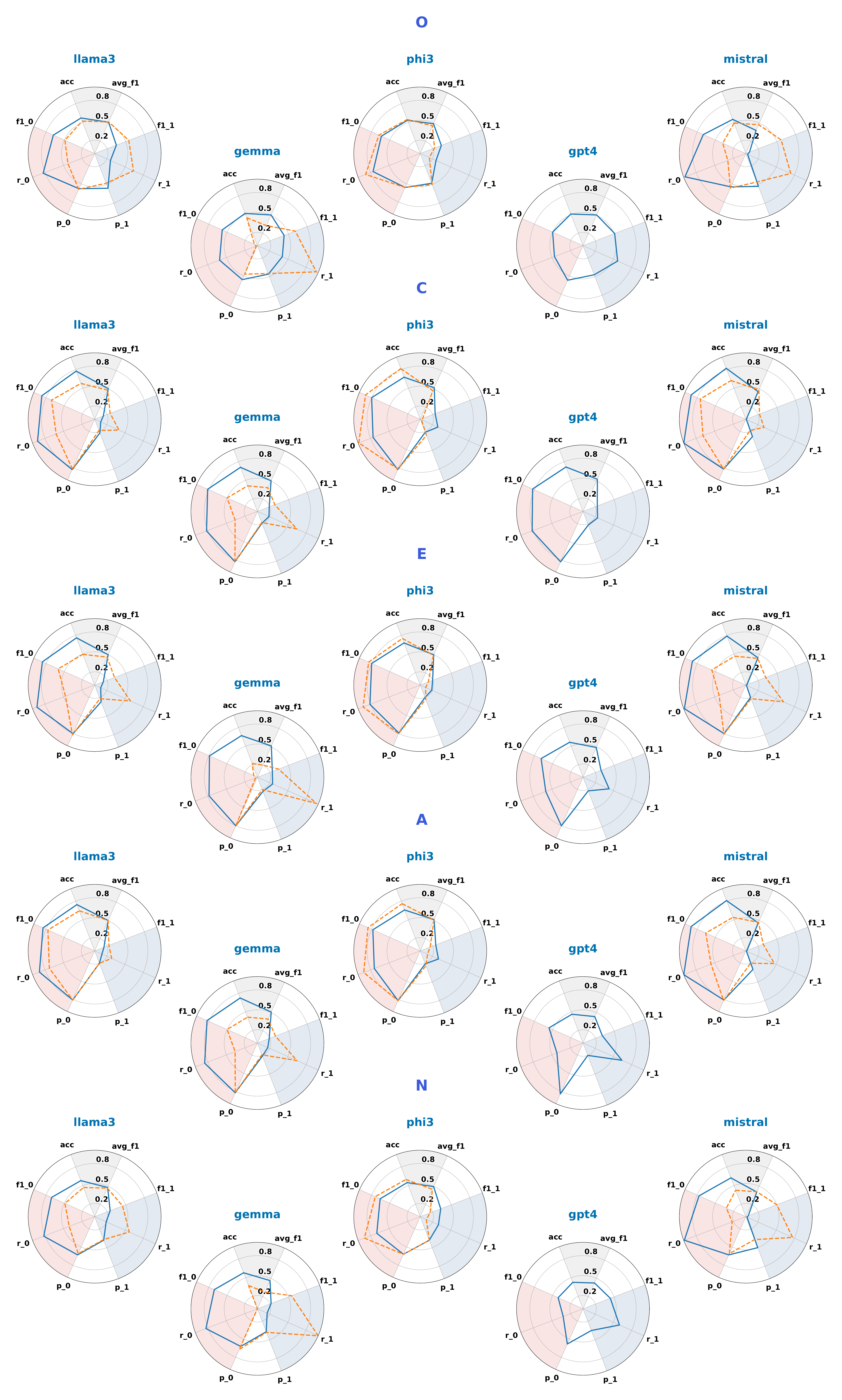}
    \caption{\textbf{Pandora} --- Description in the caption of Figure \ref{fig:prompt_radars_1}}
    \label{fig:prompt_radars_3}
\end{figure*}

In order to better understand performance changes under the complex prompt, we provide Figures \ref{fig:prompt_radars_1}-\ref{fig:prompt_radars_3} as a complement to Figures \ref{fig:prompt_delta} and \ref{fig:pandora_delta}. The simple and complex prompt are represented respectively by the solid blue line and the orange dashed one. The radar x-axis is arranged so that the upper part (grey area) highlights Accuracy and Average F1, while the left and right parts (red and blue) highlight class-wise F1, precision and recall --- for the 0 and 1 class, respectively. In this way, a trace skewed to the left indicates negatively-biased performance, vice versa if skewed to the right side. Finally, the octagon shape indicates a balanced performance.

\section{Complex prompt variables}
To ensure reproducibility, we provide all the variables used in the complex prompts in the form of a Python dictionary (Listing \ref{compelx_vars}), to be used with the complex prompt shown in Section~\ref{sec:methodology}. 

\section{Computation details}
\label{sec:computation}
We leveraged three different machines in parallel to speed up the computation, each one having a Tesla T4 GPU (14GB), with GPU memory per file limited to 50\% (peak of 7GB). The computations took about one month in total.

\begin{sidewaystable}
\centering
\caption{All experiments with class-wise F1 $\geq$ 0.5 (sorted in descending order on Accuracy, F1\_1 and F1\_0)}
\label{tab:good-table}
\resizebox{\linewidth}{!}{%
\begin{tabular}{lllllllrrrrrrrrr}
\toprule
\rowcolor{white}
 & Prompt type & Model & Trait & Test data & Support\_0 & Support\_1 & \% Class 1 & Accuracy & F1\_0 & F1\_1 & Avg F1 & Precision\_0 & Recall\_0 & Precision\_1 & Recall\_1 \\
\midrule
1 & complex & mistral:instruct & A & MYPERSONALITY & 116 & 134 & 0.536 & 0.608 & 0.600 & 0.620 & 0.610 & 0.570 & 0.620 & 0.650 & 0.600 \\
2 & simple & gpt4 & A & ESSAYS & 1143 & 1296 & 0.531 & 0.593 & 0.540 & 0.630 & 0.585 & 0.570 & 0.520 & 0.610 & 0.660 \\
3 & simple & mistral:instruct & O & ESSAYS & 1196 & 1271 & 0.515 & 0.584 & 0.520 & 0.630 & 0.575 & 0.590 & 0.470 & 0.580 & 0.690 \\
4 & complex & llama3.1:8b & A & MYPERSONALITY & 116 & 134 & 0.536 & 0.568 & 0.570 & 0.570 & 0.570 & 0.530 & 0.610 & 0.610 & 0.530 \\
5 & complex & phi3:latest & C & ESSAYS & 1214 & 1253 & 0.508 & 0.559 & 0.590 & 0.530 & 0.560 & 0.540 & 0.640 & 0.580 & 0.490 \\
6 & complex & mistral:instruct & A & ESSAYS & 1157 & 1310 & 0.531 & 0.556 & 0.570 & 0.550 & 0.560 & 0.520 & 0.620 & 0.600 & 0.500 \\
7 & complex & phi3:latest & A & ESSAYS & 1157 & 1310 & 0.531 & 0.545 & 0.510 & 0.580 & 0.545 & 0.520 & 0.500 & 0.570 & 0.580 \\
8 & simple & phi3:latest & A & MYPERSONALITY & 104 & 121 & 0.538 & 0.542 & 0.560 & 0.530 & 0.545 & 0.500 & 0.620 & 0.590 & 0.470 \\
9 & complex & llama3.1:8b & C & MYPERSONALITY & 120 & 130 & 0.520 & 0.524 & 0.520 & 0.530 & 0.525 & 0.500 & 0.540 & 0.550 & 0.510 \\
10 & simple & llama3.1:8b & A & ESSAYS & 1155 & 1308 & 0.531 & 0.523 & 0.500 & 0.540 & 0.520 & 0.490 & 0.520 & 0.550 & 0.530 \\
11 & simple & gemma:7b-instruct & O & ESSAYS & 1196 & 1271 & 0.515 & 0.518 & 0.530 & 0.510 & 0.520 & 0.500 & 0.560 & 0.540 & 0.480 \\
12 & simple & gpt4 & O & PANDORA & 7532 & 6209 & 0.452 & 0.510 & 0.500 & 0.510 & 0.505 & 0.570 & 0.460 & 0.470 & 0.570 \\
13 & complex & phi3:latest & N & MYPERSONALITY & 151 & 99 & 0.396 & 0.504 & 0.510 & 0.500 & 0.505 & 0.630 & 0.430 & 0.410 & 0.620 \\
\bottomrule
\end{tabular}
}
\end{sidewaystable}


\clearpage
\begin{sidewaystable}
\small
\centering
\vspace{0cm} 
\caption{All experiments with Accuracy and F1\_1 $\geq$ 0.5 (sorted in descending order on Accuracy and F1\_1)}
\label{tab:misleading-table}
\resizebox{\linewidth}{!}{%
\begin{tabular}{lllllllrrrrrrrrr}
\toprule
\rowcolor{white}
 & Prompt type & Model & Trait & Test data & Support\_0 & Support\_1 & \% Class 1 & Accuracy & F1\_0 & F1\_1 & Avg F1 & Precision\_0 & Recall\_0 & Precision\_1 & Recall\_1 \\
\midrule
1 & complex & llama3.1:8b & O & MYPERSONALITY & 74 & 176 & 0.704 & 0.632 & 0.220 & 0.760 & 0.490 & 0.300 & 0.180 & 0.700 & 0.820 \\
2 & simple & gpt4 & O & MYPERSONALITY & 74 & 173 & 0.700 & 0.628 & 0.280 & 0.750 & 0.515 & 0.330 & 0.240 & 0.710 & 0.790 \\
3 & complex & mistral:instruct & O & MYPERSONALITY & 74 & 176 & 0.704 & 0.620 & 0.330 & 0.740 & 0.535 & 0.340 & 0.310 & 0.720 & 0.750 \\
4 & complex & mistral:instruct & A & MYPERSONALITY & 116 & 134 & 0.536 & 0.608 & 0.600 & 0.620 & 0.610 & 0.570 & 0.620 & 0.650 & 0.600 \\
5 & simple & gpt4 & A & ESSAYS & 1143 & 1296 & 0.531 & 0.593 & 0.540 & 0.630 & 0.585 & 0.570 & 0.520 & 0.610 & 0.660 \\
6 & simple & llama3.1:8b & O & MYPERSONALITY & 72 & 175 & 0.709 & 0.587 & 0.430 & 0.680 & 0.555 & 0.360 & 0.530 & 0.760 & 0.610 \\
7 & simple & mistral:instruct & O & ESSAYS & 1196 & 1271 & 0.515 & 0.584 & 0.520 & 0.630 & 0.575 & 0.590 & 0.470 & 0.580 & 0.690 \\
8 & simple & llama3.1:8b & O & ESSAYS & 1192 & 1270 & 0.516 & 0.579 & 0.410 & 0.670 & 0.540 & 0.640 & 0.300 & 0.560 & 0.840 \\
9 & simple & mistral:instruct & N & ESSAYS & 1234 & 1233 & 0.500 & 0.573 & 0.470 & 0.640 & 0.555 & 0.620 & 0.380 & 0.550 & 0.770 \\
10 & simple & gpt4 & E & ESSAYS & 1180 & 1259 & 0.516 & 0.569 & 0.470 & 0.640 & 0.555 & 0.580 & 0.400 & 0.560 & 0.730 \\
11 & complex & llama3.1:8b & A & MYPERSONALITY & 116 & 134 & 0.536 & 0.568 & 0.570 & 0.570 & 0.570 & 0.530 & 0.610 & 0.610 & 0.530 \\
12 & complex & llama3.1:8b & A & ESSAYS & 1156 & 1308 & 0.531 & 0.560 & 0.440 & 0.640 & 0.540 & 0.550 & 0.360 & 0.570 & 0.730 \\
13 & complex & phi3:latest & C & ESSAYS & 1214 & 1253 & 0.508 & 0.559 & 0.590 & 0.530 & 0.560 & 0.540 & 0.640 & 0.580 & 0.490 \\
14 & complex & mistral:instruct & A & ESSAYS & 1157 & 1310 & 0.531 & 0.556 & 0.570 & 0.550 & 0.560 & 0.520 & 0.620 & 0.600 & 0.500 \\
15 & complex & llama3.1:8b & C & ESSAYS & 1214 & 1251 & 0.508 & 0.551 & 0.430 & 0.630 & 0.530 & 0.570 & 0.340 & 0.540 & 0.750 \\
16 & complex & mistral:instruct & O & ESSAYS & 1196 & 1271 & 0.515 & 0.548 & 0.380 & 0.640 & 0.510 & 0.570 & 0.290 & 0.540 & 0.790 \\
17 & complex & phi3:latest & A & ESSAYS & 1157 & 1310 & 0.531 & 0.545 & 0.510 & 0.580 & 0.545 & 0.520 & 0.500 & 0.570 & 0.580 \\
18 & simple & phi3:latest & O & MYPERSONALITY & 66 & 157 & 0.704 & 0.543 & 0.450 & 0.610 & 0.530 & 0.350 & 0.640 & 0.770 & 0.500 \\
19 & simple & gpt4 & A & MYPERSONALITY & 115 & 132 & 0.534 & 0.543 & 0.420 & 0.620 & 0.520 & 0.510 & 0.360 & 0.560 & 0.700 \\
20 & simple & phi3:latest & A & MYPERSONALITY & 104 & 121 & 0.538 & 0.542 & 0.560 & 0.530 & 0.545 & 0.500 & 0.620 & 0.590 & 0.470 \\
21 & complex & phi3:latest & E & ESSAYS & 1191 & 1276 & 0.517 & 0.540 & 0.280 & 0.660 & 0.470 & 0.570 & 0.190 & 0.530 & 0.870 \\
22 & simple & llama3.1:8b & C & ESSAYS & 1213 & 1248 & 0.507 & 0.538 & 0.490 & 0.580 & 0.535 & 0.540 & 0.460 & 0.540 & 0.620 \\
23 & simple & gpt4 & O & ESSAYS & 1183 & 1256 & 0.515 & 0.535 & 0.350 & 0.640 & 0.495 & 0.540 & 0.260 & 0.530 & 0.790 \\
24 & simple & phi3:latest & N & ESSAYS & 191 & 207 & 0.520 & 0.533 & 0.160 & 0.680 & 0.420 & 0.580 & 0.090 & 0.530 & 0.940 \\
25 & complex & llama3.1:8b & E & ESSAYS & 1190 & 1274 & 0.517 & 0.532 & 0.240 & 0.660 & 0.450 & 0.560 & 0.150 & 0.530 & 0.890 \\
26 & simple & phi3:latest & C & ESSAYS & 180 & 219 & 0.549 & 0.526 & 0.270 & 0.650 & 0.460 & 0.440 & 0.190 & 0.550 & 0.800 \\
27 & complex & gemma:7b-instruct & O & MYPERSONALITY & 74 & 176 & 0.704 & 0.524 & 0.420 & 0.600 & 0.510 & 0.330 & 0.580 & 0.740 & 0.500 \\
28 & complex & llama3.1:8b & C & MYPERSONALITY & 120 & 130 & 0.520 & 0.524 & 0.520 & 0.530 & 0.525 & 0.500 & 0.540 & 0.550 & 0.510 \\
29 & simple & llama3.1:8b & A & ESSAYS & 1155 & 1308 & 0.531 & 0.523 & 0.500 & 0.540 & 0.520 & 0.490 & 0.520 & 0.550 & 0.530 \\
30 & complex & llama3.1:8b & O & PANDORA & 7636 & 6365 & 0.455 & 0.523 & 0.490 & 0.550 & 0.520 & 0.590 & 0.430 & 0.480 & 0.640 \\
31 & simple & phi3:latest & A & ESSAYS & 184 & 190 & 0.508 & 0.521 & 0.290 & 0.640 & 0.465 & 0.540 & 0.200 & 0.520 & 0.830 \\
32 & simple & llama3.1:8b & N & ESSAYS & 1234 & 1233 & 0.500 & 0.521 & 0.150 & 0.670 & 0.410 & 0.660 & 0.090 & 0.510 & 0.960 \\
33 & complex & phi3:latest & O & MYPERSONALITY & 74 & 176 & 0.704 & 0.520 & 0.390 & 0.610 & 0.500 & 0.310 & 0.510 & 0.720 & 0.520 \\
34 & complex & llama3.1:8b & O & ESSAYS & 1194 & 1271 & 0.516 & 0.520 & 0.050 & 0.680 & 0.365 & 0.590 & 0.030 & 0.520 & 0.980 \\
35 & simple & gemma:7b-instruct & O & ESSAYS & 1196 & 1271 & 0.515 & 0.518 & 0.530 & 0.510 & 0.520 & 0.500 & 0.560 & 0.540 & 0.480 \\
36 & complex & phi3:latest & O & ESSAYS & 1196 & 1271 & 0.515 & 0.516 & 0.050 & 0.680 & 0.365 & 0.520 & 0.030 & 0.520 & 0.980 \\
37 & complex & gemma:7b-instruct & N & ESSAYS & 1234 & 1233 & 0.500 & 0.513 & 0.150 & 0.660 & 0.405 & 0.590 & 0.090 & 0.510 & 0.940 \\
38 & simple & llama3.1:8b & E & ESSAYS & 1187 & 1273 & 0.517 & 0.512 & 0.430 & 0.570 & 0.500 & 0.490 & 0.380 & 0.520 & 0.640 \\
39 & simple & phi3:latest & O & ESSAYS & 235 & 230 & 0.495 & 0.510 & 0.250 & 0.640 & 0.445 & 0.550 & 0.160 & 0.500 & 0.870 \\
40 & simple & gpt4 & O & PANDORA & 7532 & 6209 & 0.452 & 0.510 & 0.500 & 0.510 & 0.505 & 0.570 & 0.460 & 0.470 & 0.570 \\
41 & simple & phi3:latest & E & ESSAYS & 241 & 248 & 0.507 & 0.509 & 0.330 & 0.610 & 0.470 & 0.500 & 0.240 & 0.510 & 0.770 \\
42 & complex & phi3:latest & N & ESSAYS & 1234 & 1233 & 0.500 & 0.505 & 0.030 & 0.670 & 0.350 & 0.770 & 0.020 & 0.500 & 1.000 \\
43 & complex & mistral:instruct & N & ESSAYS & 1234 & 1233 & 0.500 & 0.505 & 0.030 & 0.670 & 0.350 & 0.810 & 0.010 & 0.500 & 1.000 \\
44 & complex & llama3.1:8b & N & ESSAYS & 1234 & 1232 & 0.500 & 0.505 & 0.020 & 0.670 & 0.345 & 0.930 & 0.010 & 0.500 & 1.000 \\
45 & complex & phi3:latest & N & MYPERSONALITY & 151 & 99 & 0.396 & 0.504 & 0.510 & 0.500 & 0.505 & 0.630 & 0.430 & 0.410 & 0.620 \\
\bottomrule
\end{tabular}
}
\end{sidewaystable}
\clearpage

\clearpage

\clearpage

\begin{lstlisting}[caption={Complex prompt strategy variables}, label={compelx_vars}]
#high-level from Costa 92 NEO-PI-R
short_trait_descr={
    "Neuroticism": "Neuroticism is the dimension underlying a broad group of traits, including anxiety, hostility, and the tendency to experience negative emotions such as guilt and sadness. Related to psychological distress and emotional instability.",
    "Extraversion": "Extraversion is the dimension underlying a broad group of traits, including sociability, activity, and the tendency to experience positive emotions such as joy and pleasure. High scorers are social, talkative, assertive, and energetic.",
    "Openness": "High-Openness individuals are imaginative and sensitive to art and beauty and have a rich and complex emotional life; they are intellectually curious, behaviorally flexible, and nondogmatic in their attitudes and values. Related to creativity, openness to novelty, and appreciation for art.",
    "Agreeableness": "Agreeableness is primarily a dimension of interpersonal behavior. High-Agreeableness individuals are trusting, sympathetic, and cooperative; low-Agreeablenness individuals are cynical, callous, and antagonistic. Involves compassion, altruism, and trust in others.",
    "Conscientiousness": "Conscientiousness is a dimension with social, vocational implications, associated with academic and vocational success. High scorers are disciplined, organized, reliable."
}


#low-level from Goldberg 90 IPIP (facets, 20-item scale)
long_trait_descr={    
    "Extraversion":{
        "positive":["Am the life of the party.",
         	"Feel comfortable around people.",
         	"Start conversations.",
         	"Talk to a lot of different people at parties.",
         	"Don't mind being the center of attention.",
         	"Make friends easily.",
         	"Take charge.",
         	"Know how to captivate people.",
         	"Feel at ease with people.",
         	"Am skilled in handling social situations."],
        "negative":[ "Don't talk a lot.",
         	"Keep in the background.",
         	"Have little to say.",
         	"Don't like to draw attention to myself.",
         	"Am quiet around strangers.",
         	"Find it difficult to approach others.",
         	"Often feel uncomfortable around others.",
         	"Bottle up my feelings.",
         	"Am a very private person.",
         	"Wait for others to lead the way."]
    },

    "Agreeableness":{
        "positive":["Am interested in people.",
         	"Sympathize with others' feelings.",
         	"Have a soft heart.",
         	"Take time out for others.",
         	"Feel others' emotions.",
         	"Make people feel at ease.",
         	"Inquire about others' well-being.",
         	"Know how to comfort others.",
         	"Love children.",
         	"Am on good terms with nearly everyone.",
         	"Have a good word for everyone.",
         	"Show my gratitude.",
         	"Think of others first.",
         	"Love to help others."],
        "negative":["Insult people.",
         	"Am not interested in other people's problems.",
         	"Feel little concern for others.",
         	"Am not really interested in others.",
         	"Am hard to get to know.",
         	"Am indifferent to the feelings of others."],
    },

    "Conscientiousness":{
        "positive":["Am always prepared.",
         	"Pay attention to details.",
         	"Get chores done right away.",
         	"Like order.",
         	"Follow a schedule.",
         	"Am exacting in my work.",
         	"Do things according to a plan.",
         	"Continue until everything is perfect.",
         	"Make plans and stick to them.",
         	"Love order and regularity.",
         	"Like to tidy up."],
        "negative":["Leave my belongings around.",
         	"Make a mess of things.",
         	"Often forget to put things back in their proper place.",
         	"Shirk my duties.",
         	"Neglect my duties.",
         	"Waste my time.",
         	"Do things in a half-way manner.",
         	"Find it difficult to get down to work.",
         	"Leave a mess in my room."]
    },
    #NEUROTICISM IS INVERTED (the original is Emotional Stability)!
    "Neuroticism":{
        "negative":["Am relaxed most of the time.",
         	"Seldom feel blue.",
         	"Am not easily bothered by things.",
         	"Rarely get irritated.",
         	"Seldom get mad."],
        "positive":["Get stressed out easily.",
         	"Worry about things.",
         	"Am easily disturbed.",
         	"Get upset easily.",
         	"Change my mood a lot.",
         	"Have frequent mood swings.",
         	"Get irritated easily.",
         	"Often feel blue.",
         	"Get angry easily.",
         	"Panic easily.",
         	"Feel threatened easily.",
         	"Get overwhelmed by emotions.",
         	"Take offense easily.",
         	"Get caught up in my problems.",
         	"Grumble about things."]
    },

    "Openness":{
        "positive":["Have a rich vocabulary.",
         	"Have a vivid imagination.",
         	"Have excellent ideas.",
         	"Am quick to understand things.",
         	"Use difficult words.",
         	"Spend time reflecting on things.",
         	"Am full of ideas.",
         	"Carry the conversation to a higher level.",
         	"Catch on to things quickly.",
         	"Can handle a lot of information.",
         	"Love to think up new ways of doing things.",
         	"Love to read challenging material.",
         	"Am good at many things."],
        "negative":["Have difficulty understanding abstract ideas."
         	"Am not interested in abstract ideas."
         	"Do not have a good imagination."
         	"Try to avoid complex people."
         	"Have difficulty imagining things."
         	"Avoid difficult reading material."
         	"Will not probe deeply into a subject."]
    }
}

#from goldberg 1990 paper, Table 1
adjectives={
    "Extraversion":{
        "positive":['jolly',
             'merry',
             'witty',
             'lively',
             'peppy',
             'talkative',
             'articulate',
             'verbose',
             'gossipy',
             'companionable',
             'social',
             'outgoin',
             'impulsive',
             'carefree',
             'playful',
             'zany',
             'mischievous',
             'rowdy',
             'loud',
             'prankish',
             'brave',
             'venturous',
             'fearless',
             'reckless',
             'active',
             'assertive',
             'dominant',
             'energetic',
             'boastful',
             'conceited',
             'egotistical',
             'affected',
             'vain',
             'chic',
             'dapper',
             'jaunt',
             'nosey',
             'snoopy',
             'indiscreet',
             'meddlesom',
             'sexy',
             'passionate',
             'sensual',
             'flirtatious'],
        "negative":['reserved',
             'lethargic',
             'vigorless',
             'apathetic',
             'cool',
             'aloof',
             'distant',
             'unsocial',
             'withdrawn',
             'quiet',
             'secretive',
             'untalkative',
             'indirect',
             'humble',
             'modest',
             'bashful',
             'meek',
             'shy',
             'joyless',
             'solemn',
             'sober',
             'morose',
             'moody',
             'utactless',
             'thoughtless',
             'unfriendly']
    },

    "Agreeableness":{
        "positive":['trustful',
             'unsuspicious',
             'unenvious',
             'democratic',
             'friendly',
             'genial',
             'cheerful',
             'generous',
             'charitable',
             'indulgent',
             'lenient',
             'conciliatory',
             'cooperative',
             'agreeable',
             'tolerant',
             'reasonable',
             'impartial',
             'unbiased',
             'patient',
             'moderate',
             'tactful',
             'polite',
             'civil',
             'kind',
             'loyal',
             'unselfish',
             'helpful',
             'sensitive',
             'affectionate',
             'warm',
             'tender',
             'sentimental',
             'moral',
             'honest',
             'just',
             'principled'],
        "negative":['sadistic',
             'vengeful',
             'cruel',
             'malicious',
             'bitter',
             'testy',
             'crabby',
             'sour',
             'surly',
             'harsh',
             'severe',
             'strict',
             'critical',
             'bossy',
             'derogatory',
             'caustic',
             'sarcastic',
             'catty',
             'negative',
             'contrary',
             'argumentative',
             'belligerent',
             'abrasive',
             'unruly',
             'aggressive',
             'biased',
             'opinionated',
             'stubborn',
             'inflexible',
             'irritable',
             'explosive',
             'wild',
             'short-tempered',
             'jealous',
             'mistrustful',
             'suspicious',
             'stingy',
             'selfish',
             'ungenerous',
             'envious',
             'scheming',
             'sly',
             'wily',
             'insincere',
             'devious']
    },

    "Conscientiousness":{
        "positive":['persistent',
             'ambitious',
             'organized',
             'thorough',
             'orderly',
             'prim',
             'tidy',
             'discreet',
             'controlled',
             'serious',
             'earnest',
             'crusading',
             'zealous',
             'moralistic',
             'prudish',
             'predictable',
             'rigid',
             'conventional',
             'rational',
             'courtly',
             'dignified',
             'genteel',
             'suave',
             'conscientious',
             'dependable',
             'prompt',
             'punctual',
             'blase',
             'urbane',
             'cultured',
             'refined',
             'formal',
             'pompous',
             'smug',
             'proud',
             'aimful',
             'calculating',
             'farseeing',
             'progressive',
             'mystical',
             'devout',
             'pious',
             'spiritual',
             'mature',
             'coy',
             'demure',
             'chaste',
             'unvoluptuous',
             'economical',
             'frugal',
             'thrifty',
             'unextravagant'],
        "negative":['messy',
             'forgetful',
             'lazy',
             'careless',
             'changeable',
             'erratic',
             'fickle',
             'absent-minded',
             'impolite',
             'impudent',
             'rude',
             'cynical',
             'nonreligious',
             'informal',
             'profane',
             'awkward',
             'unrefined',
             'earthy',
             'practical',
             'thriftless',
             'excessive',
             'self-indulgent']
    },
    #INVERTED!
    "Neuroticism":{
        "positive":['touchy',
             'careworn',
             'whiny',
             'oversensitive',
             'fearful',
             'nervous',
             'fussy',
             'unstable',
             'unconfident',
             'self-critical',
             'unpoised',
             'cowardly',
             'timid',
             'unventurous',
             'wary',
             'docile',
             'dependent',
             'submissive',
             'pliant',
             'naive',
             'gullible',
             'superstitious',
             'childlike'],
        "negative":['tough',
             'rugged',
             'unflinching',
             'wordless',
             'calm',
             'stable',
             'sedate',
             'peaceful',
             'confident',
             'independent',
             'resourceful',
             'ruthless',
             'insensitive',
             'cold',
             'stern',
             'frank',
             'blunt',
             'explicit',
             'curt',
             'terse']
    },

    "Openness":{
        "positive":['intelligent',
             'philosophical',
             'complex',
             'deep',
             'insightful',
             'clever',
             'creative',
             'curious',
             'alert',
             'perceptive',
             'logical',
             'certain',
             'informed',
             'literate',
             'studious',
             'intellectual',
             'pensive',
             'thoughtful',
             'meditative',
             'literary',
             'poetic',
             'artistic',
             'musical'],
        "negative":['simple', 'ignorant', 'dull', 'illogical', 'narrow']
    }
}
\end{lstlisting}

\end{document}